\documentclass{article}

\usepackage[nonatbib,final]{neurips}

\usepackage{color,xcolor}
\usepackage{epsfig}
\usepackage{graphicx}

\usepackage{adjustbox}
\usepackage{array}
\usepackage{booktabs}
\usepackage{colortbl}
\usepackage{float,wrapfig}
\usepackage{hhline}
\usepackage{multirow}
\usepackage{tabularx}

\usepackage{amsmath,amsfonts,amsthm,amssymb}
\usepackage{bm}
\usepackage{nicefrac}
\usepackage{microtype}

\usepackage{changepage}
\usepackage{extramarks}
\usepackage{fancyhdr}
\usepackage{lastpage}
\usepackage{setspace}
\usepackage{soul}
\usepackage{xspace}

\usepackage[pagebackref=true,breaklinks=true,colorlinks,citecolor=gray,urlcolor={purple}]{hyperref}
\usepackage[nocompress]{cite}
\usepackage{url}

\usepackage{algorithm, algorithmic}
\usepackage{enumerate}
\usepackage{todonotes}

\usepackage{etoolbox,siunitx}
\usepackage{afterpage}
\usepackage{subcaption}
\usepackage[title]{appendix}

\newcommand{\fid}{Fr\'echet Inception Distance\xspace}

\newcommand{\reffig}[1]{Figure~\ref{fig:#1}}
\newcommand{\refsec}[1]{Section~\ref{sec:#1}}
\newcommand{\refapp}[1]{Appendix~\ref{sec:#1}}
\newcommand{\reftbl}[1]{Table~\ref{tab:#1}}

\newcommand{\lblsec}[1]{\label{sec:#1}}

\newcommand{\lbltbl}[1]{\label{tab:#1}}

\newcommand{\sect}[1]{Section~\ref{#1}}

\newcommand{\eqs}[1]{Equations~(\ref{#1})}
\newcommand{\fig}[1]{Figure~\ref{#1}}
\newcommand{\figs}[1]{Figures~\ref{#1}}
\newcommand{\tab}[1]{Table~\ref{#1}}
\newcommand{\tabs}[1]{Tables~\ref{#1}}

\newcommand{\ignorethis}[1]{}

\makeatletter
\DeclareRobustCommand\onedot{\futurelet\@let@token\@onedot}
\def\@onedot{\ifx\@let@token.\else.\null\fi\xspace}

\def\eg{\emph{e.g}\onedot} 
\def\ie{\emph{i.e}\onedot} 
 
 \def\vs{\emph{vs}\onedot}
 
\def\etal{\emph{et al}\onedot}
\makeatother

\newcommand{\vv}[1]{\boldsymbol{#1}}
\newcommand{\E}{\mathop{\mathbb{E}}}
\newrobustcmd{\B}{\bfseries}
\newcommand*{\rom}[1]{\expandafter\romannumeral #1}

\definecolor{mydarkblue}{rgb}{0,0.08,1}
\definecolor{mydarkgreen}{rgb}{0.02,0.6,0.02}
\definecolor{mydarkred}{rgb}{0.8,0.02,0.02}
\definecolor{mydarkorange}{rgb}{0.40,0.2,0.02}
\definecolor{mypurple}{RGB}{111,0,255}
\definecolor{myred}{rgb}{1.0,0.0,0.0}
\definecolor{mygold}{rgb}{0.75,0.6,0.12}
\definecolor{myblue}{rgb}{0,0.2,0.8}
\definecolor{mydarkgray}{rgb}{0.66,0.66,0.66}

\newcommand{\added}[1]{#1}

\newcommand{\myparagraph}[1]{\vspace{-4pt}\paragraph{#1}}

\def\method{Differentiable Augmentation\xspace}
\def\methodshort{DiffAugment\xspace}

\title{Differentiable Augmentation \\ for Data-Efficient GAN Training}
\author{
  Shengyu Zhao \\
  IIIS, Tsinghua University and MIT\\
   \And
  Zhijian Liu \\
  MIT\\
   \And
  Ji Lin \\
  MIT\\
   \And
  Jun-Yan Zhu \\
  Adobe and CMU \\
   \And
  Song Han \\
  MIT\\
}

\begin{document}

\maketitle

\begin{abstract}

The performance of generative adversarial networks (GANs) heavily deteriorates given a limited amount of training data. This is mainly because the discriminator is memorizing the exact training set. To combat it, we propose \emph{\method} (\emph{\methodshort}), a simple method that improves the data efficiency of GANs by imposing various types of differentiable augmentations on both real and fake samples. Previous attempts to directly augment the training data manipulate the distribution of real images, yielding little benefit; \methodshort enables us to adopt the differentiable augmentation for the generated samples, effectively stabilizes training, and leads to better convergence. Experiments demonstrate consistent gains of our method over a variety of GAN architectures and loss functions for both unconditional and class-conditional generation. With \methodshort, we achieve a state-of-the-art FID of 6.80 with an IS of 100.8 on ImageNet 128$\times$128 and 2-4$\times$ reductions of FID given 1,000 images on FFHQ and LSUN. Furthermore, with only 20\% training data, we can match the top performance on CIFAR-10 and CIFAR-100. Finally, our method can generate high-fidelity images using only 100 images without pre-training, while being on par with existing transfer learning algorithms. Code is available at \url{https://github.com/mit-han-lab/data-efficient-gans}.

\end{abstract}
\section{Introduction}

\added{Big data has enabled deep learning algorithms achieve rapid advancements. In particular, state-of-the-art generative adversarial networks (GANs)~\cite{goodfellow2014GAN}} are able to generate high-fidelity natural images of diverse categories~\cite{brock2018BigGAN,karras2019StyleGAN2}. Many computer vision and graphics applications have been enabled~\cite{zhu2017unpaired,wang2017fast,park2019semantic}. However, this success comes at the cost of a tremendous amount of computation and data. Recently, researchers have proposed promising techniques to improve the \textit{computational efficiency} of model inference~\cite{li2020gan,shu2019co}, while the \textit{data efficiency} remains to be a fundamental challenge.

\begin{figure}[t]
\begin{center}
\includegraphics[width=1.0\linewidth]{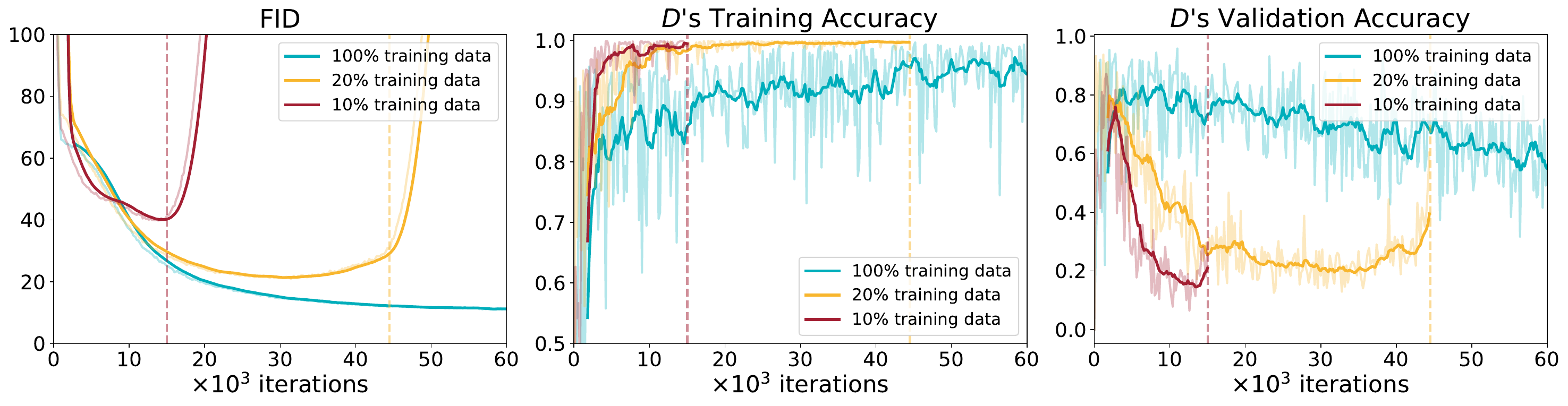}
\end{center}
\vspace{-8pt}
\caption{\textbf{BigGAN heavily deteriorates given a limited amount of data.} {\em left}: With 10\% of CIFAR-10 data, FID increases shortly after the training starts, and the model then collapses (red curve). {\em middle}: the training accuracy of the discriminator $D$ quickly saturates. {\em right}: the validation accuracy of $D$ dramatically falls, indicating that $D$ has memorized the exact training set and fails to generalize.%
}
\label{fig:cifar_curves}
\end{figure}

GANs heavily rely on vast quantities of diverse and high-quality training examples. To name a few, the FFHQ dataset~\cite{karras2019StyleGAN} contains 70,000 selective post-processed high-resolution images of human faces; the ImageNet dataset~\cite{deng2009imagenet} annotates more than a million of images with various object categories. Collecting such large-scale datasets requires \emph{months or even years} of considerable human efforts along with prohibitive annotation costs. In some cases, it is not even possible to have that many examples, \eg, images of rare species or photos of a specific person or landmark. Thus, it is of critical importance to eliminate the need of immense datasets for GAN training. However, reducing the amount of training data results in drastic degradation in the performance. For example in \fig{fig:cifar_curves}, given only 10\% or 20\% of the CIFAR-10 data, the training accuracy of the discriminator saturates quickly (to nearly 100\%); however, its validation accuracy keeps decreasing (to lower than 30\%), suggesting that the discriminator is simply memorizing the entire training set. This severe over-fitting problem disrupts the training dynamics and leads to degraded image quality.

A widely-used strategy to reduce overfitting in image classification is data augmentation~\cite{simard2003best,wan2013regularization,krizhevsky2012imagenet}, which can increase the diversity of training data without collecting new samples. Transformations such as cropping, flipping, scaling, color jittering~\cite{krizhevsky2012imagenet}, and region masking (Cutout)~\cite{devries2017Cutout} are commonly-used augmentations for vision models. However, applying data augmentation to GANs is fundamentally different. If the transformation is only added to the real images, the generator would be encouraged to match the distribution of the \textit{augmented} images. As a consequence, the outputs suffer from distribution shift and the introduced artifacts (\eg, a region being masked, unnatural color, see \fig{fig:augment-aug}). Alternatively, we can augment both the real and generated images when training the discriminator; however, this would break the subtle balance between the generator and discriminator, leading to poor convergence as they are optimizing completely different objectives (see \fig{fig:augment-baug}).

\begin{figure}[t]
\begin{center}
\includegraphics[width=1.0\linewidth]{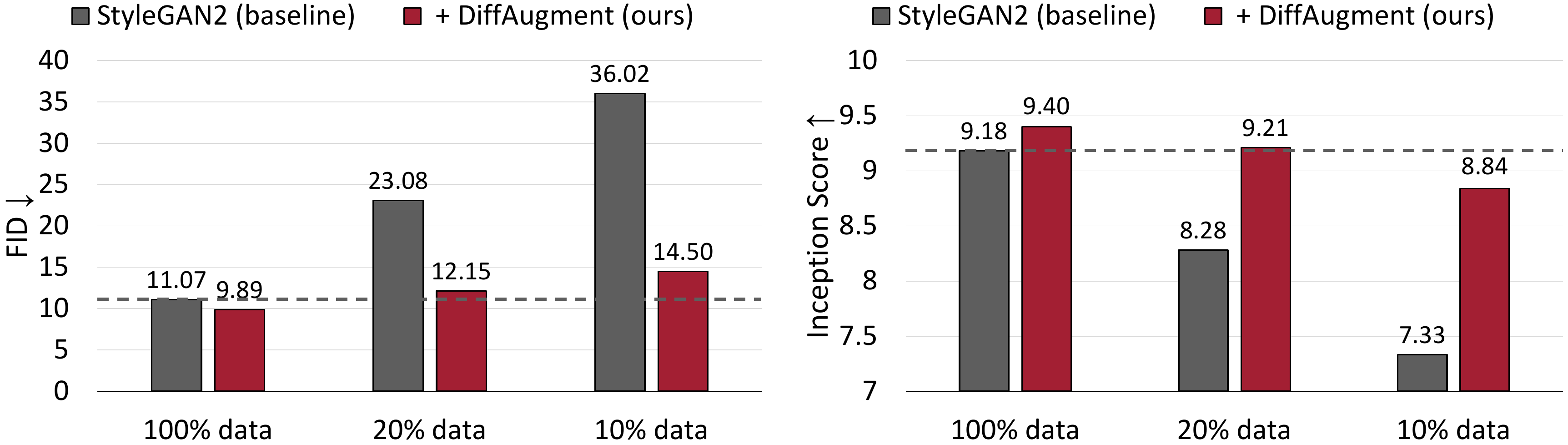}
\end{center}
\vspace{-6pt}
\caption{\textbf{Unconditional generation results on CIFAR-10.} StyleGAN2's performance drastically degrades given less training data. With \methodshort, we are able to roughly match its FID and outperform its Inception Score (IS) using only 20\% training data.
FID and IS are measured using 10k samples; the validation set is used as the reference distribution for FID calculation.
}
\label{fig:cifar_results}
\end{figure}

\begin{figure}[t]
\begin{center}
\includegraphics[width=1.0\linewidth]{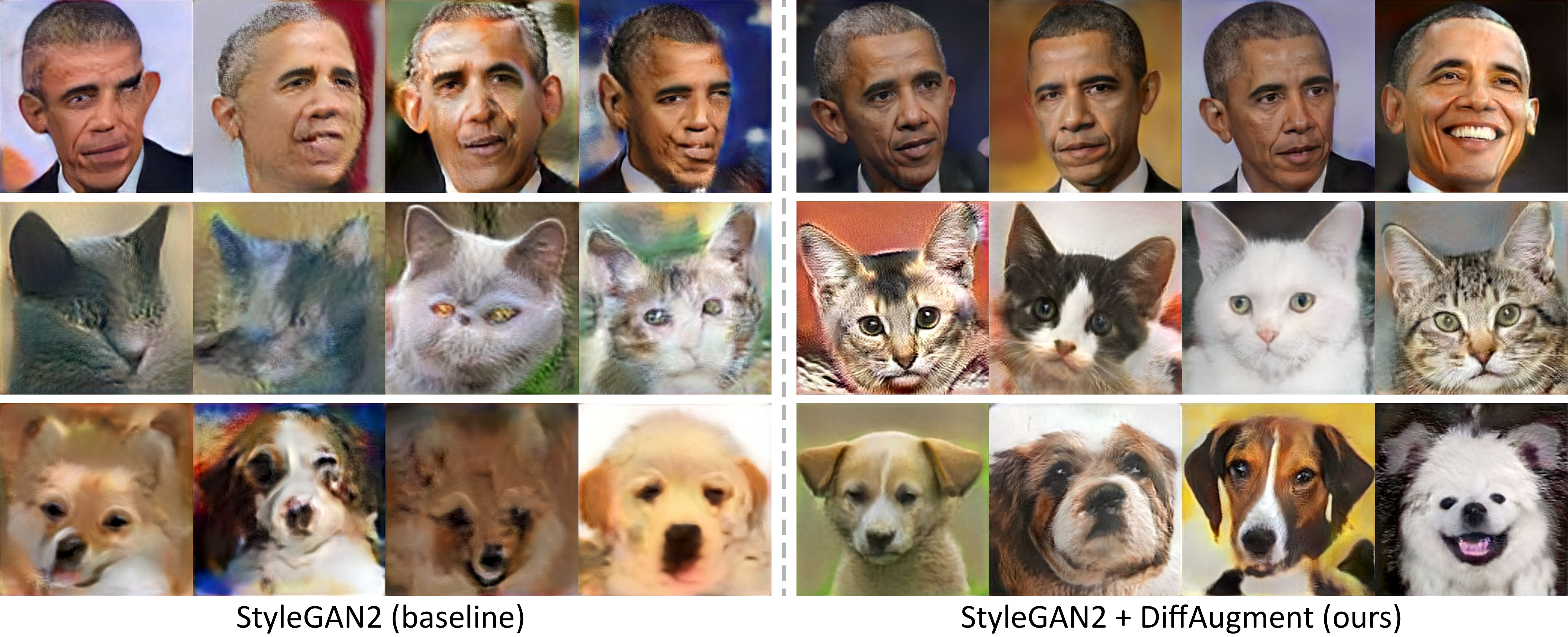}
\end{center}
\vspace{-6pt}
\caption{\textbf{Low-shot generation without pre-training.} Our method can generate high-fidelity images using only 100 Obama portraits (top) from our collected 100-shot datasets, 160 cats (middle) or 389 dogs (bottom) from the AnimalFace dataset~\cite{si2011AnimalFace} at 256$\times$256 resolution. See \fig{fig:few_shot-interp} for the interpolation results; nearest neighbor tests are provided in \refapp{app:fewshot} (\figs{fig:nearest_neighbors}-\ref{fig:nearest_neighbors_lpips}).}
\label{fig:fewshot_results}
\end{figure}

To combat it, we introduce a simple but effective method, \emph{\methodshort}, which applies the same differentiable augmentation to both real and fake images for both generator and discriminator training.
It enables the gradients to be propagated through the augmentation back to the generator, regularizes the discriminator without manipulating the target distribution, and maintains the balance of training dynamics.
Experiments on a variety of GAN architectures and datasets consistently demonstrate the effectiveness of our method. With \methodshort, we improve BigGAN and achieve a \fid (FID) of 6.80 with an Inception Score (IS) of 100.8 on ImageNet 128$\times$128 without the truncation trick~\cite{brock2018BigGAN} \added{and reduce the StyleGAN2 baseline's FID by 2-4$\times$ given 1,000 images on the FFHQ and LSUN datasets}. We also match the top performance on CIFAR-10 and CIFAR-100 using only 20\% training data (see \fig{fig:cifar_results} and \fig{fig:cifar100_results}).
Furthermore, our method can generate high-quality images with only 100 examples (see \fig{fig:fewshot_results}). Without any pre-training, we achieve competitive performance with existing transfer learning algorithms \added{that used to require tens of thousands of training images.}


\section{Related Work}

\myparagraph{Generative Adversarial Networks.}
Following the pioneering work of GAN~\cite{goodfellow2014GAN}, researchers have explored different ways to improve its performance and training stability. Recent efforts are centered on more stable objective functions~\cite{arjovsky2017WGAN,mao2017least,gulrajani2017WGANGP,mescheder2018R1,salimans2016improved}, more advanced architectures~\cite{radford2015unsupervised,miyato2018SN,miyato2018cgans,zhang2019SAGAN}, and better training strategy~\cite{denton2015deep,zhang2017stackgan,karras2018progressive,liu2020diverse}.
As a result, both the visual fidelity and diversity of generated images have increased significantly. For example, BigGAN~\cite{brock2018BigGAN} is able to synthesize natural images with a wide range of object classes at high resolution, and StyleGAN~\cite{karras2019StyleGAN,karras2019StyleGAN2} can produce photorealistic face portraits with large varieties, often indistinguishable from natural photos. However, the above work paid less attention to the data efficiency aspect. A recent attempt~\cite{luvcic2019fewer,chen2019self} leverages semi- and self-supervised learning to reduce the amount of human annotation required for training. In this paper, we study a more challenging scenario where both data and labels are limited.

\myparagraph{Regularization for GANs.} GAN training often requires additional regularization as they are highly unstable. To stabilize the training dynamics, researchers have proposed several techniques including the instance noise~\cite{sonderby2016noise}, Jensen-Shannon regularization~\cite{roth2017JSR}, gradient penalties~\cite{gulrajani2017WGANGP,mescheder2018R1}, spectral normalization~\cite{miyato2018SN}, adversarial defense regularization~\cite{zhou2018fool}, and consistency regularization~\cite{zhang2020CR}. All of these regularization techniques implicitly or explicitly penalize sudden changes in the discriminator's output within a local region of the input. In this paper, we provide a different perspective, data augmentation, and we encourage the discriminator to perform well under different types of augmentation. In \refsec{expr}, we show that our method is complementary to the regularization techniques in practice.

\myparagraph{Data Augmentation.}
Many deep learning models adopt label-preserving transformations to reduce overfitting: \eg, color jittering~\cite{krizhevsky2012imagenet}, region masking~\cite{devries2017Cutout}, flipping, rotation, cropping~\cite{wan2013regularization,krizhevsky2012imagenet}, data mixing~\cite{zhang2018mixup}, and local and affine distortion~\cite{simard2003best}. Recently, AutoML~\cite{stanley2002evolving,zoph2017neural} has been used to explore adaptive augmentation policies for a given dataset and task~\cite{cubuk2019autoaugment,lim2019fast,cubuk2019randaugment}. However, applying data augmentation to generative models, such as GANs, remains an open question. Different from the classifier training where the label is invariant to transformations of the input, the goal of generative models is to learn the data distribution itself. Directly applying augmentation would inevitably alter the distribution. We present a simple strategy to circumvent the above concern. Concurrent with our work, several methods~\cite{karras2020ADA,zhao2020image,tran2020towards} independently proposed data augmentation for training GANs. We urge the readers to check out their work for more details.
\section{Method}

Generative adversarial networks (GANs) aim to model the distribution of a target dataset via a generator $G$ and a discriminator $D$. The generator $G$ maps an input latent vector $\vv z$, typically drawn from a Gaussian distribution, to its output $G(\vv z)$. The discriminator $D$ learns to distinguish generated samples $G(\vv z)$ from real observations $\vv x$. The standard GANs training algorithm alternately optimizes the discriminator's loss $L_D$ and the generator's loss $L_G$ given loss functions $f_D$ and $f_G$:
\begin{align}
    L_D &= \E\nolimits_{\vv x \sim p_{\text{data}}(\vv x)}[f_D(-D(\vv x))] + \E\nolimits_{\vv z \sim p(\vv z)}[f_D(D(G(\vv z)))], \label{eq:baselineD} \\
    L_G &= \E\nolimits_{\vv z \sim p(\vv z)}[f_G(-D(G(\vv z)))]. \label{eq:baslineG}
\end{align}
Here, different loss functions can be used, such as the non-saturating loss~\cite{goodfellow2014GAN}, where $f_D(x) = f_G(x) = \log\left(1 + e^x\right)$, and the hinge loss~\cite{miyato2018SN}, where $f_D(x) = \max(0, 1 + x)$ and $f_G(x) = x$.

Despite extensive ongoing efforts of better GAN architectures and loss functions, a fundamental challenge still exists:  the discriminator tends to \emph{memorize} the observations as the training progresses. An overfitted discriminator penalizes any generated samples other than the exact training data points, provides uninformative gradients due to poor generalization, and usually leads to training instability.

\begin{figure}[t]
\begin{center}
\includegraphics[width=1.0\linewidth]{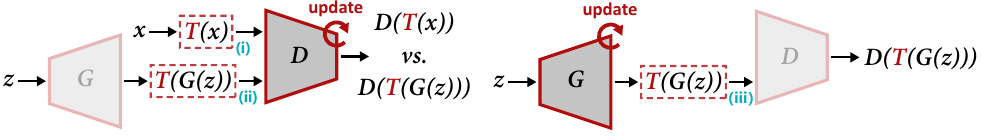}
\end{center}
\vspace{-10pt}
\caption{\textbf{Overview of \methodshort} for updating $D$ (left) and $G$ (right). \methodshort applies the augmentation $T$ to both the real samples $\vv x$ and the generated output $G(\vv z)$. When we update $G$, gradients need to be back-propagated through $T$, which requires $T$ to be differentiable w.r.t.\@ the input.}
\label{fig:method}
\end{figure}
\begin{table}[t]
\setlength{\tabcolsep}{4.8pt}
\begin{center}
\small
\sisetup{
table-number-alignment=center,
detect-weight=true,
mode=text
}
\begin{tabular}{lccc*{3}{S[table-figures-integer=1,table-figures-decimal=2]S[table-figures-integer=3,table-figures-decimal=2]}}
\toprule
\multirow{2}{*}{Method} & \multicolumn{3}{c}{Where $T$?} & \multicolumn{2}{c}{Color + Transl.\@ + Cutout} & \multicolumn{2}{c}{Transl.\@ + Cutout} & \multicolumn{2}{c}{Translation} \\
\cmidrule(lr){2-4}
\cmidrule(lr){5-6}
\cmidrule(lr){7-8}
\cmidrule(lr){9-10}
{} & \scriptsize{\textbf{(\rom{1})}} & \scriptsize{\textbf{(\rom{2})}} & \scriptsize{\textbf{(\rom{3})}} & {IS} & {FID} & {IS} & {FID} & {IS} & {FID} \\
\midrule
BigGAN (baseline) & & & &9.06   &9.59   &9.06   &9.59   &9.06   &9.59   \\
\midrule
Aug.\@ reals only & \checkmark & & &5.94	&49.38	&6.51	&37.95	&8.40	&19.16  \\
Aug.\@ reals + fakes ($D$ only)  & \checkmark & \checkmark & &3.00	&126.96	&3.76	&114.14	&3.50	&100.13 \\
\methodshort ($D$ + $G$, ours) & \checkmark & \checkmark & \checkmark &\B 9.25&\B 8.59&\B 9.16&\B 8.70&\B 9.07&\B 9.04\\
\bottomrule
\end{tabular}
\end{center}
\caption{\textbf{\methodshort vs.\@ vanilla augmentation strategies} on CIFAR-10 with 100\% training data. ``Augment reals only'' applies augmentation $T$ to (\rom{1}) only (see \reffig{method}) and corresponds to \eqs{eq:augD}-(\ref{eq:augG}); ``Augment $D$ only'' applies $T$ to both reals~(\rom{1}) and fakes~(\rom{2}), but not $G$~(\rom{3}), and corresponds to \eqs{eq:nondiffD}-(\ref{eq:nondiffG}); ``\methodshort'' applies $T$ to reals~(\rom{1}), fakes~(\rom{2}), and $G$~(\rom{3}). (\rom{3}) requires $T$ to be differentiable since gradients should be back-propagated through $T$ to $G$. \methodshort corresponds to \eqs{eq:diffaugD}-(\ref{eq:diffaugG}).
IS and FID are measured using 10k samples; the validation set is the reference distribution. We select the snapshot with the best FID for each method. Results are averaged over 5 evaluation runs; all standard deviations are less than 1\% relatively.}
\label{tab:augment}
\vspace{-8pt}
\end{table}
\begin{figure}[t]
\begin{center}
\subcaptionbox{``Augment reals only'': the same augmentation \\ artifacts appear on the generated images.  \label{fig:augment-aug}}{
\includegraphics[width=0.48\linewidth]{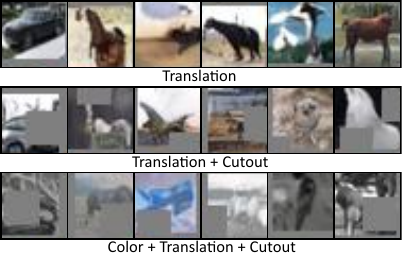}
\vspace{2pt}
}
\subcaptionbox{``Augment $D$ only'': the unbalanced optimization between $G$ and $D$ cripples training. \label{fig:augment-baug}}{
\includegraphics[width=0.48\linewidth]{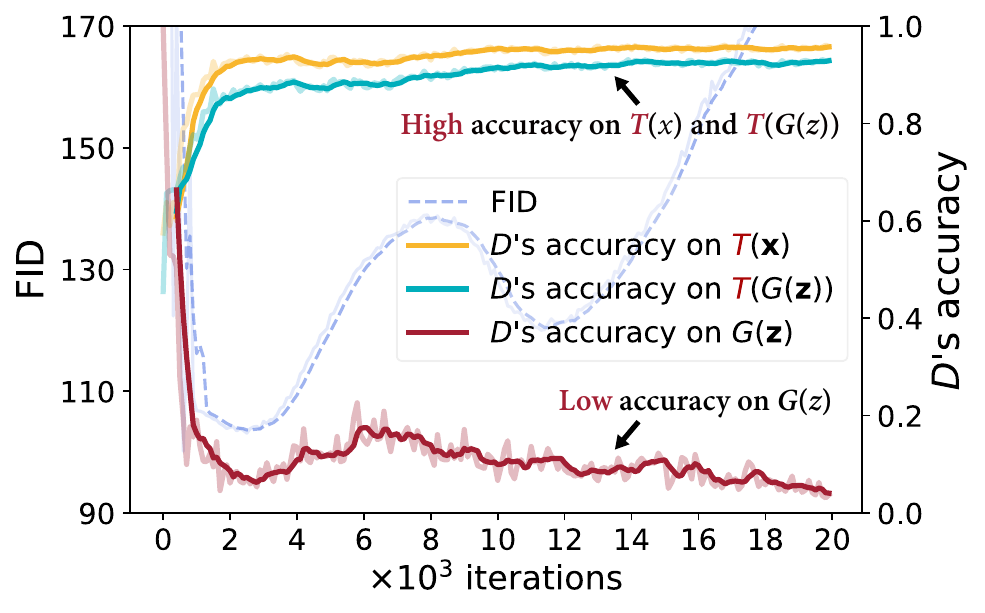}
\vspace{1pt}
}
\end{center}
\vspace{-4pt}
\caption{\textbf{Understanding why vanilla augmentation strategies fail:} (a) ``Augment reals only'' mimics the same data distortion as introduced by the augmentations, \eg, the translation padding, the Cutout square, and the color artifacts; (b) ``Augment $D$ only'' diverges because of the unbalanced optimization --- $D$ perfectly classifies the augmented images (both $T(\vv x)$ and $T(G(\vv z))$ but barely recognizes $G(\vv z)$ (i.e., fake images without augmentation) from which $G$ receives gradients.}
\label{fig:augment}
\vspace{-4pt}
\end{figure}

\myparagraph{Challenge: Discriminator Overfitting.}
Here we analyze the performance of BigGAN~\cite{brock2018BigGAN} with different amounts of data on CIFAR-10. 
As plotted in \fig{fig:cifar_curves}, even given 100\% data, the gap between the discriminator's training and validation accuracy keeps increasing, suggesting that the discriminator is simply memorizing the training images. This happens not only on limited data but also on the large-scale ImageNet dataset, as observed by Brock~\etal~\cite{brock2018BigGAN}. BigGAN already adopts Spectral Normalization~\cite{miyato2018SN}, a widely-used regularization technique for both generator and discriminator architectures, but still suffers from severe overfitting.%

\subsection{Revisiting Data Augmentation}

Data augmentation is a commonly-used strategy to reduce overfitting in many recognition tasks --- it has an irreplaceable role and can also be applied in conjunction with other regularization techniques: \eg, weight decay. We have shown that the discriminator suffers from a similar overfitting problem as the binary classifier. However, data augmentation is seldom used in the GAN literature compared to the explicit regularizations on the discriminator~\cite{gulrajani2017WGANGP,mescheder2018R1,miyato2018SN}. In fact, a recent work~\cite{zhang2020CR} observes that directly applying data augmentation to GANs does not improve the baseline.
So, we would like to ask the questions: what prevents us from simply applying data augmentation to GANs?
Why is augmenting GANs not as effective as augmenting classifiers?

\myparagraph{Augment reals only.} The most straightforward way of augmenting GANs would be directly applying augmentation $T$ to the real observations $\vv x$, which we call ``Augment reals only'':
\begin{align}
    L_D &= \E\nolimits_{\vv x \sim p_{\text{data}}(\vv x)}[f_D(-D(\textcolor{mydarkred}{T}(\vv x)))] + \E\nolimits_{\vv z \sim p(\vv z)}[f_D(D(G(\vv z)))], \label{eq:augD} \\
    L_G &= \E\nolimits_{\vv z \sim p(\vv z)}[f_G(-D(G(\vv z)))]. \label{eq:augG}
\end{align}
However, ``Augment reals only'' deviates from the original purpose of generative modeling, as the model is now learning a different data distribution of $T(\vv x)$ instead of $\vv x$. This prevents us from applying any augmentation that significantly alters the distribution of the real images. The choices that meet this requirement, although strongly dependent on the specific dataset, can only be horizontal flips in most cases. We find that applying random horizontal flips does increase the performance moderately, and we use it in all our experiments to make our baselines stronger. We demonstrate the side effects of enforcing stronger augmentations quantitatively in \tab{tab:augment} and qualitatively in \fig{fig:augment-aug}. As expected, the model learns to produce unwanted color and geometric distortion (\eg, unnatural color, cutout holes) as introduced by these augmentations, resulting in a significantly worse performance (see ``Augment reals only'' in \tab{tab:augment}).

\myparagraph{Augment $D$ only.}

Previously, ``Augment reals only'' applies one-sided augmentation to the real samples, and hence the convergence can be achieved only if the generated distribution matches the manipulated real distribution. From the discriminator's perspective, it may be tempting to augment both real and fake samples when we update $D$:
\begin{align}
    L_D &= \E\nolimits_{\vv x \sim p_{\text{data}}(\vv x)}[f_D(-D(\textcolor{mydarkred}{T}(\vv x)))] + \E\nolimits_{\vv z \sim p(\vv z)}[f_D(D(\textcolor{mydarkred}{T}(G(\vv z))))], \label{eq:nondiffD}\\
    L_G &= \E\nolimits_{\vv z \sim p(\vv z)}[f_G(-D(G(\vv z)))]. \label{eq:nondiffG}
\end{align}
Here, the same function $T$ is applied to both real samples $\vv x$ and fake samples $G(\vv z)$. If the generator successfully models the distribution of $\vv x$, $T(G(\vv z))$ and $T(\vv x)$ should be indistinguishable to the discriminator as well as $G(\vv z)$ and $\vv x$. 
However, this strategy leads to even worse results (see ``Augment $D$ only'' in \tab{tab:augment}). \fig{fig:augment-baug} plots the training dynamics of ``Augment $D$ only'' with \emph{Translation} applied. Although $D$ classifies the augmented images (both $T(G(\vv z))$ and $T(\vv x)$) perfectly with an accuracy of above 90\%, it fails to recognize $G(\vv z)$, the generated images without augmentation, with an accuracy of lower than 10\%. As a result, the generator completely fools the discriminator by $G(\vv z)$ and cannot obtain useful information from the discriminator. This suggests that any attempts that break the delicate balance between the generator $G$ and discriminator $D$ are prone to failure.

\subsection{\method for GANs}

\begin{figure}[t]
\begin{center}
\includegraphics[width=1.0\linewidth]{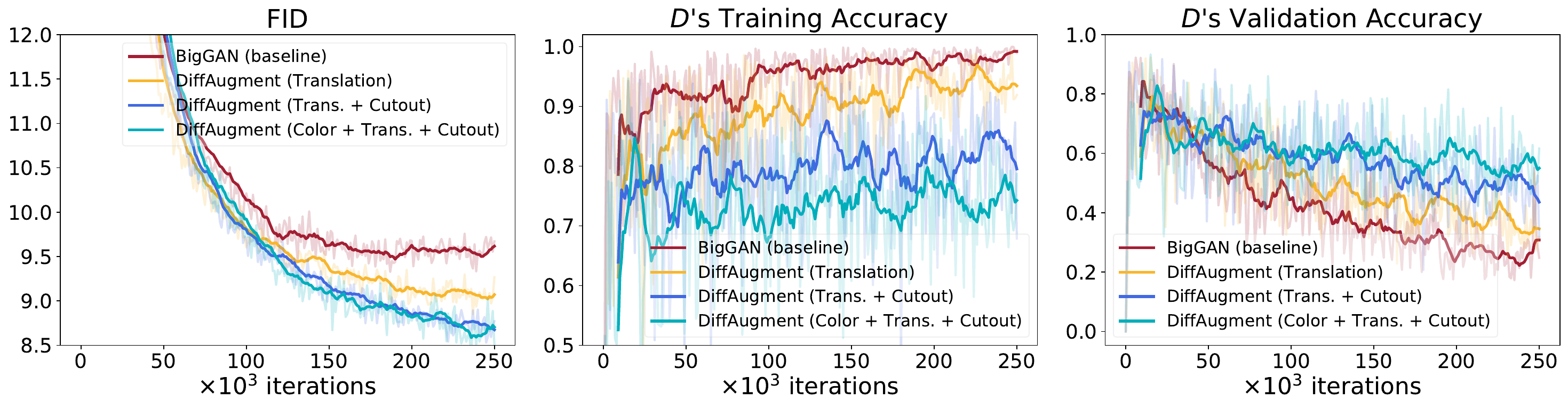}
\end{center}
\vspace{-8pt}
\caption{\textbf{Analysis of different types of \methodshort} on CIFAR-10 with 100\% training data. A stronger \methodshort can dramatically reduce the gap between the discriminator's training accuracy (middle) and validation accuracy (right), leading to a better convergence (left).}
\label{fig:cifar10-aug}
\vspace{-4pt}
\end{figure}

The failure of ``Augment reals only'' motivates us to augment both \emph{real} and \emph{fake} samples, while the failure of ``Augment $D$ only'' warns us that the \emph{generator} should not neglect the augmented samples. Therefore, to  propagate gradients through the augmented samples to $G$, the augmentation $T$ must be \emph{differentiable} as depicted in \fig{fig:method}. We call this \emph{\method} (\emph{\methodshort}):
\begin{align}
    L_D &= \E\nolimits_{\vv x \sim p_{\text{data}}}(\vv x)[f_D(-D(\textcolor{mydarkred}{T}(\vv x)))] + \E\nolimits_{\vv z \sim p(\vv z)}[f_D(D(\textcolor{mydarkred}{T}(G(\vv z))))], \label{eq:diffaugD} \\
    L_G &= \E\nolimits_{\vv z \sim p(\vv z)}[f_G(-D(\textcolor{mydarkred}{T}(G(\vv z))))]. \label{eq:diffaugG}
\end{align}
Note that $T$ is required to be the same (random) function but not necessarily the same random seed across the three places illustrated in~\fig{fig:method}.
We demonstrate the effectiveness of \methodshort using three simple choices of transformations and its composition, throughout the paper: \emph{Translation} (within $[-1/8, 1/8]$ of the image size, padded with zeros), \emph{Cutout}~\cite{devries2017Cutout} (masking with a random square of half image size), and \emph{Color} (including random brightness within $[-0.5, 0.5]$, contrast within $[0.5, 1.5]$, and saturation within $[0, 2]$). As shown in \tab{tab:augment}, BigGAN can be improved using the simple \emph{Translation} policy and further boosted using a composition of \emph{Cutout} and \emph{Translation}; it is also robust to the strongest policy when \emph{Color} is used in combined. \fig{fig:cifar10-aug} analyzes that stronger \methodshort policies generally maintain a higher discriminator's validation accuracy at the cost of a lower training accuracy, alleviate the overfitting problem, and eventually achieve better convergence. 

\section{Experiments}
\lblsec{expr}
We conduct extensive experiments on ImageNet~\cite{deng2009imagenet}, CIFAR-10~\cite{cifar10}, CIFAR-100, FFHQ~\cite{karras2019StyleGAN}, and LSUN-Cat~\cite{yu2015LSUN} based on the leading class-conditional BigGAN~\cite{brock2018BigGAN} and unconditional StyleGAN2~\cite{karras2019StyleGAN2}. We use the common evaluation metrics \fid (FID)~\cite{heusel2017FID} and Inception Score (IS)~\cite{salimans2016improved}. In addition, we apply our method to low-shot generation both with and without pre-training in \sect{sec:few-shot}. Finally, we perform analysis studies in \refsec{analysis}.

\begin{table}[t]
\begin{center}
\small
\sisetup{
table-number-alignment=center,
separate-uncertainty=true,
table-figures-uncertainty=1,
detect-weight=true,
mode=text
}
\begin{tabular}{lS[table-figures-integer=3,table-figures-decimal=1]S[table-figures-integer=1,table-figures-decimal=2]S[table-figures-integer=2,table-figures-decimal=1]S[table-figures-integer=1,table-figures-decimal=2]S[table-figures-integer=2,table-figures-decimal=1]S[table-figures-integer=1,table-figures-decimal=2]}
\toprule
\multirow{2}{*}{Method} & \multicolumn{2}{c}{100\% training data} & \multicolumn{2}{c}{50\% training data} & \multicolumn{2}{c}{25\% training data} \\
\cmidrule(lr){2-3}
\cmidrule(lr){4-5}
\cmidrule(lr){6-7}
{} & {IS} & {FID} & {IS} & {FID} & {IS} & {FID} \\
\midrule
BigGAN~\cite{brock2018BigGAN}          &94.5 \pm 0.4      &7.62 \pm 0.02	    &89.9 \pm 0.2	    &9.64 \pm 0.04	    &46.5 \pm 0.4     &25.37 \pm 0.07      \\
+ \methodshort   &\B 100.8 \pm 0.2  &\B 6.80 \pm 0.02	&\B 91.9 \pm 0.5	&\B 8.88 \pm 0.06	&\B 74.2 \pm 0.5  &\B 13.28 \pm 0.07   \\
\bottomrule
\end{tabular}
\end{center}
\caption{\textbf{ImageNet} 128$\times$128 results without the truncation trick~\cite{brock2018BigGAN}. IS and FID are measured using 50k samples; the validation set is used as the reference distribution for FID. We select the snapshot with the best FID for each method. We report means and standard deviations over 3 evaluation runs.}
\label{tab:imagenet}
\vspace{-8pt}
\end{table}

\begin{table}[t]
\setlength{\tabcolsep}{9.8pt}
\begin{center}
\small
\sisetup{
table-number-alignment=center,
detect-weight=true,
mode=text
}
\begin{tabular}{l*{1}{S[table-figures-integer=1,table-figures-decimal=2]}*{7}{S[table-figures-integer=2,table-figures-decimal=2]}}
\toprule
\multirow{2}{*}{Method} & \multicolumn{4}{c}{FFHQ} & \multicolumn{4}{c}{LSUN-Cat} \\
\cmidrule(lr){2-5}
\cmidrule(lr){6-9}
{} & {30k} & {10k} & {5k} & {1k} & {30k} & {10k} & {5k} & {1k} \\
\midrule
ADA~\cite{karras2020ADA} & 5.46 & 8.13 & 10.96 & \B 21.29	&10.50	&13.13	&16.95 &43.25\\
\midrule
StyleGAN2~\cite{karras2019StyleGAN2} & 6.16 & 14.75 & 26.60 & 62.16	&10.12	&17.93	&34.69 &182.85 \\
 + DiffAugment & \B 5.05 & \B 7.86 & \B 10.45 & 25.66	&\B 9.68	&\B 12.07	&\B 16.11 &\B 42.26 \\
\bottomrule
\end{tabular}
\end{center}
\caption{\textbf{FFHQ and LSUN-Cat} results with 1k, 5k, 10k, and 30k training samples. With the fixed \emph{Color + Translation + Cutout} \methodshort, our method improves the StyleGAN2 baseline and is on par with a concurrent work ADA~\cite{karras2020ADA}. FID is measured using 50k generated samples; the full training set is used as the reference distribution. We select the snapshot with the best FID for each method. Results are averaged over 5 evaluation runs; all standard deviations are less than 1\% relatively.}
\lbltbl{ffhq}
\vspace{-10pt}
\end{table}
\begin{table}[t]
\setlength{\tabcolsep}{8.5pt}
\begin{center}
\small
\sisetup{
table-number-alignment=center,
detect-weight=true,
mode=text
}
\begin{tabular}{l*{2}{S[table-figures-integer=2,table-figures-decimal=2]}S[table-figures-integer=1,table-figures-decimal=2]*{2}{S[table-figures-integer=2,table-figures-decimal=2]}S[table-figures-integer=2,table-figures-decimal=2]}
\toprule
\multirow{2}{*}{Method} & \multicolumn{3}{c}{CIFAR-10} & \multicolumn{3}{c}{CIFAR-100} \\
\cmidrule(lr){2-4}
\cmidrule(lr){5-7}
{} & {100\% data} & {20\% data} & {10\% data} & {100\% data} & {20\% data} & {10\% data} \\
\midrule
BigGAN~\cite{brock2018BigGAN} &9.59	&21.58	&39.78	&12.87	&33.11	&66.71 \\
+ \methodshort &\B 8.70	&\B 14.04	&\B 22.40	&\B 12.00	&\B 22.14	&\B 33.70 \\
\midrule
CR-BigGAN~\cite{zhang2020CR} &9.06	&20.62	&37.45	&11.26	&36.91	&47.16 \\
+ \methodshort &\B 8.49	&\B 12.84	&\B 18.70	&\B 11.25	&\B 20.28	&\B 26.90 \\
\midrule
StyleGAN2~\cite{karras2019StyleGAN2} &11.07	&23.08	&36.02	&16.54	&32.30	&45.87 \\
+ \methodshort &\B 9.89	&\B 12.15	&\B 14.50	&\B 15.22	&\B 16.65	&\B 20.75 \\
\bottomrule
\end{tabular}
\end{center}
\caption{\textbf{CIFAR-10 and CIFAR-100} results. We select the snapshot with the best FID for each method. Results are averaged over 5 evaluation runs; all standard deviations are less than 1\% relatively. We use 10k samples and the validation set as the reference distribution for FID calculation, as done in prior work~\cite{zhang2020CR}. Concurrent works~\cite{karras2020ADA,ho2020denoising} use a different protocol: 50k samples and the training set as the reference distribution. If we adopt this evaluation protocol, our BigGAN + DiffAugment achieves an FID of 4.61, CR-BigGAN + DiffAugment achieves an FID of 4.30, and StyleGAN2 + DiffAugment achieves an FID of 5.79.}
\label{tab:cifar}
\vspace{-8pt}
\end{table}
\begin{table}[t]
\setlength{\tabcolsep}{10pt}
\begin{center}
\small
\sisetup{
table-number-alignment=center,
detect-weight=true,
mode=text
}
\begin{tabular}{lc*{3}{S[table-figures-integer=2,table-figures-decimal=2]}*{2}{S[table-figures-integer=3,table-figures-decimal=2]}}
\toprule
\multirow{2}{*}{Method} & \multirow{2}{*}{Pre-training?} & \multicolumn{3}{c}{100-shot} & \multicolumn{2}{c}{AnimalFace~\cite{si2011AnimalFace}} \\
\cmidrule(lr){3-5}
\cmidrule(lr){6-7}
{} & {} & {Obama} & {Grumpy cat} & {Panda} & {Cat} & {Dog} \\
\midrule
Scale/shift~\cite{noguchi2019SB}	&Yes	&50.72	&34.20	&21.38	&54.83	&83.04 \\
MineGAN~\cite{wang2019MineGAN}	&Yes	&50.63	&34.54	&14.84	&54.45	&93.03 \\
\midrule
TransferGAN~\cite{wang2018TransferGAN}	&Yes	&48.73	&34.06	&23.20	&52.61	&82.38 \\
 + DiffAugment	&Yes	&\B 39.85	&\B 29.77	&\B 17.12	&\B 49.10	&\B 65.57 \\
\midrule
FreezeD~\cite{mo2020FreezeD}	&Yes	&41.87	&31.22	&17.95	&47.70	&70.46 \\
 + DiffAugment	&Yes	&\B 35.75	&\B 29.34	&\B 14.50	&\B 46.07	&\B 61.03 \\
\midrule
\midrule
StyleGAN2~\cite{karras2019StyleGAN2}	&No	&80.20	&48.90	&34.27	&71.71	&130.19 \\
 + DiffAugment	&No	&\B 46.87	&\B 27.08	&\B 12.06	&\B 42.44	&\B 58.85 \\
\bottomrule
\end{tabular}
\end{center}
\caption{\textbf{Low-shot generation} results. \added{With only \textbf{100} (Obama, Grumpy cat, Panda), \textbf{160} (Cat), or \textbf{389} (Dog) training images, our method is on par with the transfer learning algorithms that are pre-trained with \textbf{70,000} images.} FID is measured using 5k generated samples; the training set is the reference distribution. We select the snapshot with the best FID for each method.}
\lbltbl{few_shot}
\vspace{-10pt}
\end{table}
\begin{figure}[t]
\begin{center}
\includegraphics[width=1.0\linewidth]{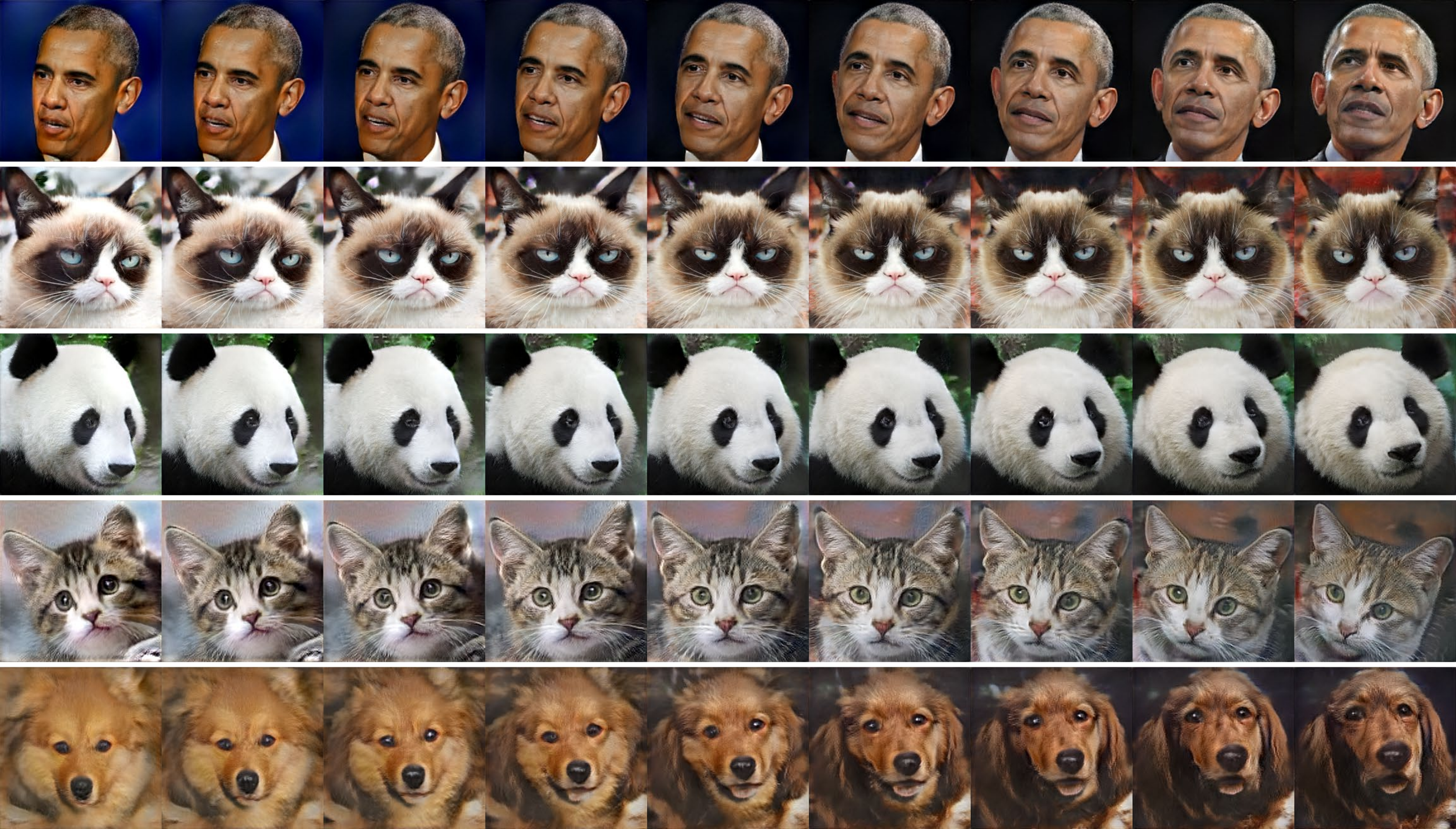}
\end{center}
\vspace{-6pt}
\caption{\textbf{Style space interpolation} of our method for low-shot generation without pre-training. The smooth interpolation results suggest little overfitting of our method even given small datasets.
}
\label{fig:few_shot-interp}
\vspace{-4pt}
\end{figure}

\begin{figure}[t]
\begin{center}
\subcaptionbox{Impact of model size. \label{fig:cifar10_1-analysis-size}}{
\includegraphics[width=0.48\linewidth]{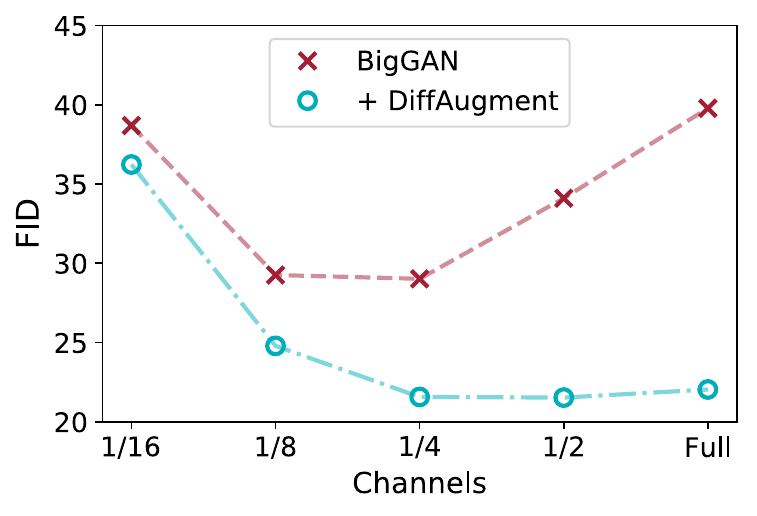}
\vspace{-2pt}
}
\subcaptionbox{Impact of $R_1$ regularization $\gamma$. \label{fig:cifar10_1-analysis-reg}}{
\includegraphics[width=0.48\linewidth]{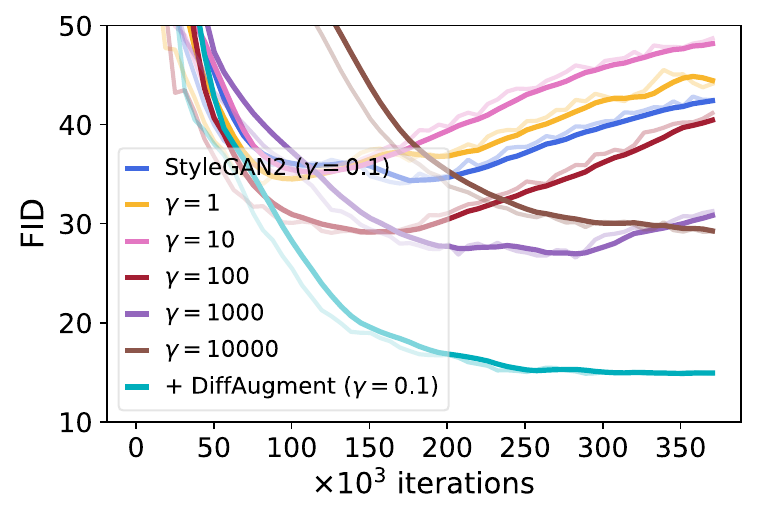}
\vspace{-2pt}
}
\end{center}
\vspace{-6pt}
\caption{\textbf{Analysis of smaller models or stronger regularization} on CIFAR-10 with 10\% training data. (a) Smaller models reduce overfitting for the BigGAN baseline, while our method dominates its performance at all model capacities. (b) Over a wide sweep of the $R_1$ regularization $\gamma$ for the baseline StyleGAN2, its best FID (26.87) is still much worse than ours (14.50).}
\label{fig:cifar10_1-analysis}
\end{figure}

\subsection{ImageNet}
\lblsec{imagenet}
We follow the top-performing model BigGAN~\cite{brock2018BigGAN} on ImageNet dataset at 128$\times$128 resolution. Additionally, we augment real images with random horizontal flips, yielding the best reimplementation of BigGAN to our knowledge (FID: ours 7.6 \vs 8.7 in the original paper~\cite{brock2018BigGAN}). We use the simple \emph{Translation} \methodshort for all the data percentage settings. In \tab{tab:imagenet}, our method achieves significant gains especially under the 25\% data setting, in which the baseline model undergoes an early collapse, and advances the state-of-the-art FID and IS with 100\% data available.%

\subsection{FFHQ and LSUN-Cat}

We further experiment with StyleGAN2~\cite{karras2019StyleGAN2} on the FFHQ portrait dataset~\cite{karras2019StyleGAN} and the LSUN-Cat dataset~\cite{yu2015LSUN} at 256$\times$256 resolution. We investigate different limited data settings, with 1k, 5k, 10k, and 30k training images available. We apply the strongest \emph{Color + Translation + Cutout} \methodshort to all the StyleGAN2 baselines without any hyperparameter changes. The real images are also augmented with random horizontal flips as commonly applied in StyleGAN2~\cite{karras2019StyleGAN2}. Results are shown in \tab{tab:ffhq}. Our performance gains are considerable under all the data percentage settings. Moreover, with the fixed policies used in \methodshort, our performance is on par with ADA~\cite{karras2020ADA}, a concurrent work based on the adaptive augmentation strategy.

\subsection{CIFAR-10 and CIFAR-100}

We experiment on the class-conditional BigGAN~\cite{brock2018BigGAN} and CR-BigGAN~\cite{zhang2020CR} and unconditional StyleGAN2~\cite{karras2019StyleGAN2} models. For a fair comparison, we also augment real images with random horizontal flips for all the baselines. The baseline models already adopt advanced regularization techniques, including Spectral Normalization~\cite{miyato2018SN}, Consistency Regularization~\cite{zhang2020CR}, and $R_1$ regularization~\cite{mescheder2018R1}; however, none of them achieves satisfactory results under the 10\% data setting. For \methodshort, we adopt \emph{Translation + Cutout} for the BigGAN models, \emph{Color + Cutout} for StyleGAN2 with 100\% data, and \emph{Color + Translation + Cutout} for StyleGAN2 with 10\% or 20\% data. As summarized in \tab{tab:cifar}, our method improves all the baselines independently of the baseline architectures, regularizations, and loss functions (hinge loss in BigGAN and non-saturating loss in StyleGAN2) without any hyperparameter changes. We refer the readers to the appendix (\tabs{tab:cifar10}-\ref{tab:cifar100}) for the complete tables with IS. The improvements are considerable especially when limited data is available. This is, to our knowledge, the new state of the art on CIFAR-10 and CIFAR-100 for both class-conditional and unconditional generation under all the 10\%, 20\%, and 100\% data settings.

\subsection{Low-Shot Generation} 
\lblsec{few-shot}

For a certain person, an object, or a landmark, it is often tedious, if not completely impossible, to collect a large-scale dataset. To address this, researchers recently exploit few-shot learning~\cite{fei2006one,lake2015human} in the setting of image generation. Wang~\etal~\cite{wang2018TransferGAN} use fine-tuning to transfer the knowledge of models pre-trained on external large-scale datasets. Several works propose to fine-tune only part of the model~\cite{noguchi2019SB,wang2019MineGAN,mo2020FreezeD}. %
Below, we show that our method not only produces competitive results without using external datasets or models but also is orthogonal to the existing transfer learning methods.

We replicate the recent transfer learning algorithms~\cite{mo2020FreezeD,noguchi2019SB,wang2019MineGAN,wang2018TransferGAN} using the same codebase as Mo~\etal~\cite{mo2020FreezeD} on their datasets (AnimalFace~\cite{si2011AnimalFace} with 160 cats and 389 dogs), based on the pre-trained StyleGAN model from the FFHQ face dataset~\cite{karras2019StyleGAN}. To further demonstrate the data efficiency, we collect the 100-shot Obama, grumpy cat, and panda datasets, and train the StyleGAN2 model on each dataset using only 100 images without pre-training. For \methodshort, we adopt \emph{Color + Translation + Cutout} for StyleGAN2, \emph{Color + Cutout} for both the vanilla fine-tuning algorithm TransferGAN~\cite{wang2018TransferGAN} and FreezeD~\cite{mo2020FreezeD} that freezes the first several layers of the discriminator. \tab{tab:few_shot} shows that \methodshort achieves consistent gains independently of the training algorithm on all the datasets. Without any pre-training, we still achieve results on par with the existing transfer learning algorithms \added{that require tens of thousands of images}, with an exception on the 100-shot Obama dataset where pre-training with human faces clearly leads to better generalization. See \fig{fig:fewshot_results} and the appendix (\figs{fig:few_shot-cat}-\ref{fig:few_shot-panda}) for qualitative comparisons. While there might be a concern that the generator is likely to overfit the tiny datasets (\ie, generating identical training images), \fig{fig:few_shot-interp} suggests little overfitting of our method via linear interpolation in the style space~\cite{karras2019StyleGAN} (see also \fig{fig:fewshot_interp_suppl} in the appendix); please refer to the appendix (\figs{fig:nearest_neighbors}-\ref{fig:nearest_neighbors_lpips}) for the nearest neighbor tests.

\subsection{Analysis} \lblsec{analysis}

Below, we investigate whether smaller model or stronger regularization would similarly reduce overfitting and whether \methodshort still helps. Finally, we analyze additional choices of \methodshort.

\myparagraph{Model Size Matters?} We reduce the model capacity of BigGAN by progressively halving the number of channels for both $G$ and $D$. As plotted in \fig{fig:cifar10_1-analysis-size}, the baseline heavily overfits on CIFAR-10 with 10\% training data when using the full model and achieves a minimum FID of 29.02 at $1/4$ channels. However, it is surpassed by our method over all model capacities. With $1/4$ channels, our model achieves a significantly better FID of 21.57, while the gap is monotonically increasing as the model becomes larger. We refer the readers to the appendix (\fig{fig:cifar10_1-analysis-IS}) for the IS plot.

\myparagraph{Stronger Regularization Matters?}

As StyleGAN2 adopts the $R_1$ regularization~\cite{mescheder2018R1} to stabilize training, we increase its strength from $\gamma = 0.1$ to up to $10^4$ and plot the FID curves in \fig{fig:cifar10_1-analysis-reg}. While we initially find that $\gamma = 0.1$ works best under the 100\% data setting, the choice of $\gamma = 10^3$ boosts its performance from 34.05 to 26.87 under the 10\% data setting. When $\gamma = 10^4$, within 750k iterations, we only observe a minimum FID of 29.14 at 440k iteration and the performance deteriorates after that. However, its best FID is still \textbf{1.8$\times$} worse than ours (with the default $\gamma = 0.1$). This shows that \methodshort is more effective compared to explicitly regularizing the discriminator.

\begin{wrapfigure}{r}{0.5\linewidth}
\vspace{-22pt}
\begin{center}
\includegraphics[width=1.0\linewidth]{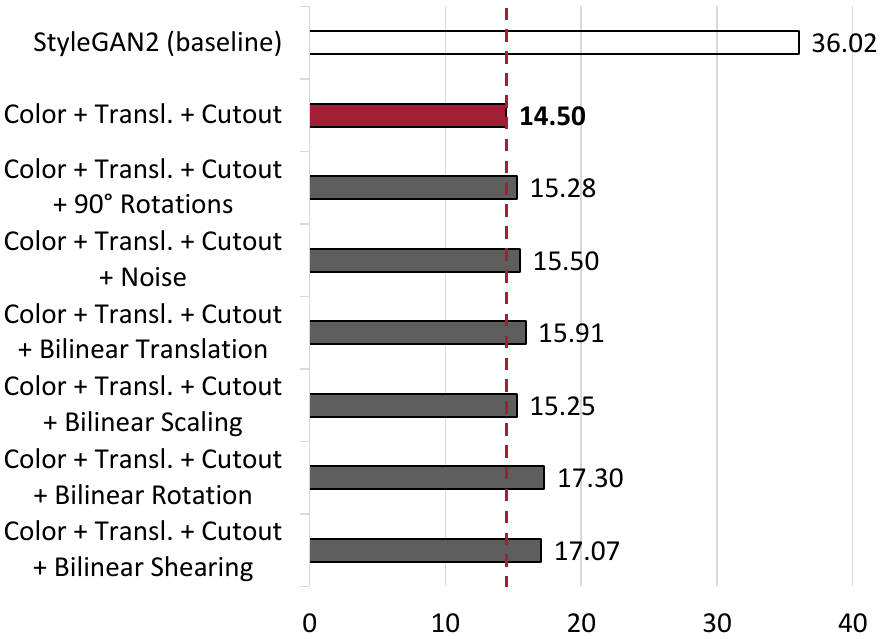}
\end{center}
\vspace{-8pt}
\caption{Various types of \methodshort consistently outperform the baseline. We report StyleGAN2's FID on CIFAR-10 with 10\% training data.}
\label{fig:aug-analysis}
\vspace{-20pt}
\end{wrapfigure}
\myparagraph{Choice of \methodshort Matters?}

\added{We investigate additional choices of \methodshort in \fig{fig:aug-analysis}, including random \ang{90} rotations (\{\ang{-90}, \ang{0}, \ang{90}\} each with $1/3$ probability), Gaussian noise (with a standard deviation of $0.1$), and general geometry transformations that involve bilinear interpolation, such as bilinear translation (within $[-0.25, 0.25]$), bilinear scaling (within $[0.75, 1.25]$), bilinear rotation (within $[\ang{-30}, \ang{30}]$), and bilinear shearing (within $[-0.25, 0.25]$). While all these policies consistently outperform the baseline, we find that the \emph{Color + Translation + Cutout} \methodshort is especially effective. The simplicity also makes it easier to deploy.}

\section{Conclusion}

We present \emph{\methodshort} for data-efficient GAN training. \methodshort reveals valuable observations that augmenting both real and fake samples effectively prevents the discriminator from over-fitting, and that the augmentation must be differentiable to enable both generator and discriminator training. Extensive experiments consistently demonstrate its benefits with different network architectures (StyleGAN2 and BigGAN), supervision settings, and objective functions, \added{across multiple datasets (ImageNet, CIFAR, FFHQ, LSUN, and 100-shot datasets)}. Our method is especially effective when limited data is available. Our \href{https://github.com/mit-han-lab/data-efficient-gans}{code, datasets, and models} are available for future comparisons.

\section*{Broader Impact}

In this paper, we investigate GANs from the data efficiency perspective, aiming to make generative modeling accessible to more people (e.g., visual artists and novice users) and research fields who have no access to abundant data. In the real-world scenarios, there could be various reasons that lead to limited amount of data available, such as rare incidents, privacy concerns, and historical visual data~\cite{ginosar2015century}. DiffAugment provides a promising way to alleviate the above issues and make AI more accessible to everyone.

\section*{Acknowledgments}
We thank NSF Career Award \#1943349, MIT-IBM Watson AI Lab, Google, Adobe, and Sony for supporting this research. Research supported with Cloud TPUs from Google's TensorFlow Research Cloud (TFRC). We thank William S. Peebles and Yijun Li for helpful comments.

{\small
\bibliographystyle{ieee}
\bibliography{main}
}

\begin{appendices}

\section{Hyperparameters and Training Details}

\subsection{ImageNet Experiments}
\lblsec{app:imagenet}

The Compare GAN codebase\footnote{\url{https://github.com/google/compare_gan}} suffices to replicate BigGAN's FID on ImageNet dataset at 128$\times$128 resolution but has some small differences to the original paper~\cite{brock2018BigGAN}. First, the codebase uses a learning rate of $10^{-4}$ for $G$ and $5 \times 10^{-4}$ for $D$. Second, it processes the raw images into 128$\times$128 resolution with random scaling and random cropping. Since we find that random cropping leads to a worse IS, we process the raw images with random scaling and center cropping instead. We additionally augment the images with random horizontal flips, yielding the best re-implementation of BigGAN to our knowledge. With \methodshort, we find that $D$'s learning rate of $5 \times 10^{-4}$ often makes $D$'s loss stuck at a high level, so we reduce $D$'s learning rate to $4 \times 10^{-4}$ for the 100\% data setting and $2 \times 10^{-4}$ for the 10\% and 20\% data settings. However, we note that the baseline model does not benefit from this reduced learning rate: if we reduce $D$'s learning rate from $5 \times 10^{-4}$ to $2 \times 10^{-4}$ under the 50\% data setting, its performance degrades from an FID/IS of 9.64/89.9 to 10.79/75.7. All the models achieve the best FID within 200k iterations and deteriorate after that, taking up to 3 days on a TPU v2/v3 Pod with 128 cores.

\begin{figure}[t]
\begin{center}
\includegraphics[width=1.0\linewidth]{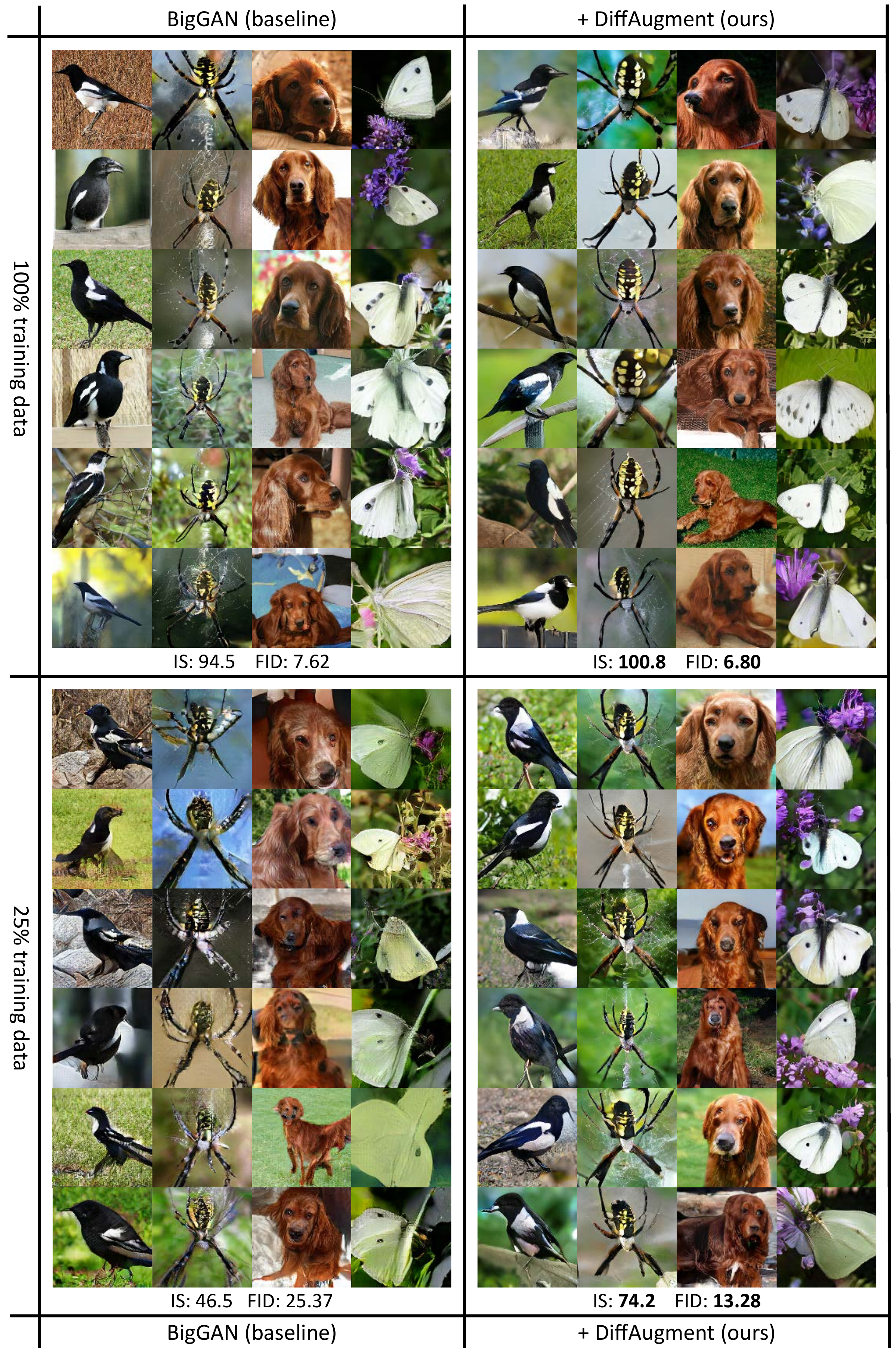}
\end{center}
\caption{Qualitative comparison on \textbf{ImageNet} 128$\times$128 without the truncation trick~\cite{brock2018BigGAN}.}
\label{fig:imagenet_comparison}
\end{figure}

See \fig{fig:imagenet_comparison} for a qualitative comparison between BigGAN and BigGAN + \methodshort. Our method improves the image quality of the samples in both 25\% and 100\% data settings. The visual difference is more clear under the 25\% data setting. 

\myparagraph{Notes on CR-BigGAN~\cite{zhang2020CR}.} CR-BigGAN~\cite{zhang2020CR} reports an FID of 6.66, which is slightly better than ours 6.80 (BigGAN + \methodshort) with 100\% data. However, the code and pre-trained models of CR-BigGAN~\cite{zhang2020CR} are not available, while its IS is not reported either. Our reimplemented CR-BigGAN only achieves an FID of 7.95 with an IS of 82.0, even worse than the baseline BigGAN. Nevertheless, our CIFAR experiments suggest the potential of applying \methodshort on top of CR.

\afterpage{
\clearpage
\begin{figure}[t]
\begin{center}
\includegraphics[width=1.0\linewidth]{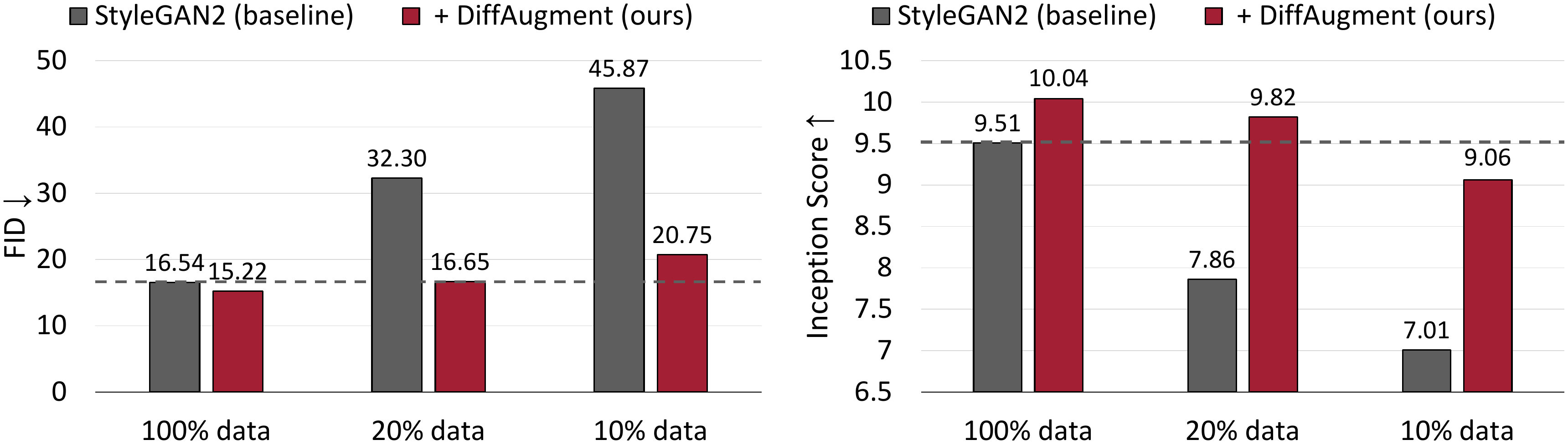}
\end{center}
\vspace{-6pt}
\caption{Unconditional generation results on \textbf{CIFAR-100}. We are able to roughly match StyleGAN2's FID and outperform its IS using only 20\% training data.
}
\label{fig:cifar100_results}
\vspace{-4pt}
\end{figure}
\begin{figure}[t]
\begin{center}
\includegraphics[width=0.48\linewidth]{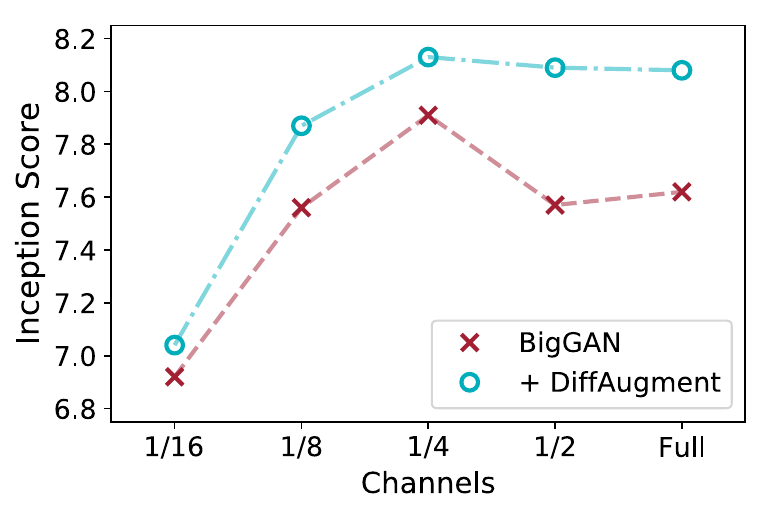}
\includegraphics[width=0.48\linewidth]{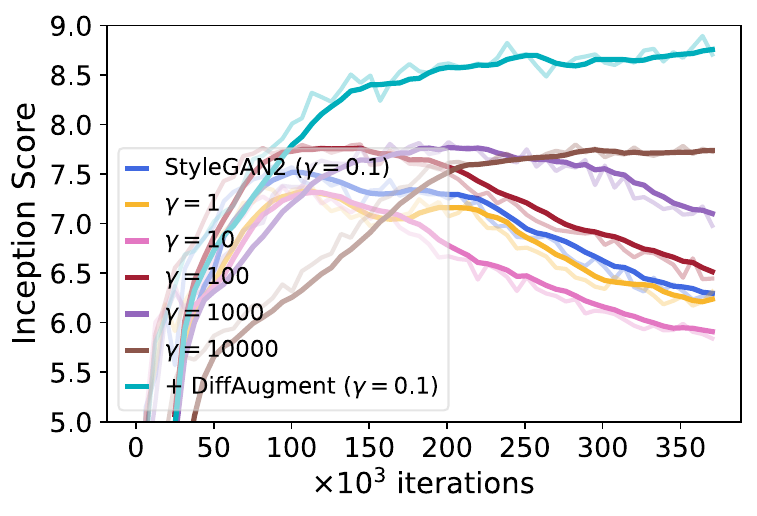}
\end{center}
\vspace{-10pt}
\caption{\textbf{Analysis of smaller models or stronger regularization} on CIFAR-10 with \textbf{10\%} training data. \emph{left}: Smaller models reduce overfitting for the BigGAN baseline, while our method outperforms it at all the model capacities. \emph{right}: Over a wide sweep of the $R_1$ regularization $\gamma$ for the baseline StyleGAN2, its best IS (7.75) is still \textbf{12\%} worse than ours (8.84).}
\label{fig:cifar10_1-analysis-IS}
\end{figure}
\begin{table}[t]
\setlength{\tabcolsep}{13pt}
\begin{center}
\sisetup{
table-number-alignment=center,
detect-weight=true,
mode=text
}
\begin{tabular}{l*{2}{S[table-figures-integer=1,table-figures-decimal=2]S[table-figures-integer=2,table-figures-decimal=2]}*{2}{S[table-figures-integer=1,table-figures-decimal=2]}}
\toprule
\multirow{2}{*}{Method} & \multicolumn{2}{c}{100\% training data} & \multicolumn{2}{c}{20\% training data} & \multicolumn{2}{c}{10\% training data} \\
\cmidrule(lr){2-3}
\cmidrule(lr){4-5}
\cmidrule(lr){6-7}
{} & {IS} & {FID} & {IS} & {FID} & {IS} & {FID} \\
\midrule
BigGAN~\cite{brock2018BigGAN} &9.06	&9.59	&8.41	&21.58	&7.62	&39.78 \\
+ \methodshort &\B 9.16	&\B 8.70	&\B 8.65	&\B 14.04	&\B 8.09	&\B 22.40 \\
\midrule
CR-BigGAN~\cite{zhang2020CR} &\B 9.20	&9.06	&8.43	&20.62	&7.66	&37.45 \\
+ \methodshort &9.17	&\B 8.49	&\B 8.61	&\B 12.84	&\B 8.49	&\B 18.70 \\
\midrule
StyleGAN2~\cite{karras2019StyleGAN2} &9.18	&11.07	&8.28	&23.08	&7.33	&36.02 \\
+ \methodshort &\B 9.40	&\B 9.89	&\B 9.21	&\B 12.15	&\B 8.84	&\B 14.50 \\
\bottomrule
\end{tabular}
\end{center}
\caption{\textbf{CIFAR-10 results.} IS and FID are measured using 10k samples; the validation set is the reference distribution for FID calculation. We select the snapshot with the best FID for each method. Results are averaged over 5 evaluation runs; all standard deviations are less than 1\% relatively.}
\label{tab:cifar10}
\vspace{-8pt}
\end{table}
\begin{table}[t]
\setlength{\tabcolsep}{13pt}
\begin{center}
\sisetup{
table-number-alignment=center,
detect-weight=true,
mode=text
}
\begin{tabular}{l*{2}{S[table-figures-integer=1,table-figures-decimal=2]S[table-figures-integer=2,table-figures-decimal=2]}*{2}{S[table-figures-integer=2,table-figures-decimal=2]}}
\toprule
\multirow{2}{*}{Method} & \multicolumn{2}{c}{100\% training data} & \multicolumn{2}{c}{20\% training data} & \multicolumn{2}{c}{10\% training data} \\
\cmidrule(lr){2-3}
\cmidrule(lr){4-5}
\cmidrule(lr){6-7}
{} & {IS} & {FID} & {IS} & {FID} & {IS} & {FID} \\
\midrule
BigGAN~\cite{brock2018BigGAN}	&\B 10.92	&12.87	&9.11	&33.11	&5.94	&66.71 \\
+ \methodshort	&10.66	&\B 12.00	&\B 9.47	&\B 22.14	&\B 8.38	&\B 33.70 \\
\midrule
CR-BigGAN~\cite{zhang2020CR}	&\B 10.95	&11.26	&8.44	&36.91	&7.91	&47.16 \\
+ \methodshort	&10.81	&\B 11.25	&\B 9.12	&\B 20.28	&\B 8.70	&\B 26.90 \\
\midrule
StyleGAN2~\cite{karras2019StyleGAN2}	&9.51	&16.54	&7.86	&32.30	&7.01	&45.87 \\
+ \methodshort	&\B 10.04	&\B 15.22	&\B 9.82	&\B 16.65	&\B 9.06	&\B 20.75 \\
\bottomrule
\end{tabular}
\end{center}
\caption{\textbf{CIFAR-100 results.} IS and FID are measured using 10k samples; the validation set is the reference distribution for FID calculation. We select the snapshot with the best FID for each method. Results are averaged over 5 evaluation runs; all standard deviations are less than 1\% relatively.}
\label{tab:cifar100}
\vspace{-8pt}
\end{table}
\clearpage
}

\subsection{FFHQ and LSUN-Cat Experiments}

We use the official TensorFlow implementation of StyleGAN2\footnote{\url{https://github.com/NVlabs/stylegan2}} and the default network configuration at 256$\times$256 resolution with an $R_1$ regularization $\gamma$ of $1$, but without the path length regularization and the lazy regularization since they do not improve FID~\cite{karras2019StyleGAN2}. The number of feature maps at shallow layers (64$\times$64 resolution and above) is halved to match the architecture of ADA~\cite{karras2020ADA}. All the models in our experiments are augmented with random horizontal flips, trained on 8 GPUs with a maximum training length of 25,000k images.

See \fig{fig:ffhq_comparison}-\ref{fig:lsun_cat_comparison} for qualitative comparisons between StyleGAN2 and StyleGAN2 + \methodshort. Our method considerably improves the image quality with limited data available.

\subsection{CIFAR-10 and CIFAR-100 Experiments}
\lblsec{app:cifar}
We replicate BigGAN and CR-BigGAN baselines on CIFAR using the PyTorch implementation\footnote{\url{https://github.com/ajbrock/BigGAN-PyTorch}}. All hyperparameters are kept unchanged from the default CIFAR-10 configuration, including the batch size (50), the number of $D$ steps (4) per $G$ step, and a learning rate of $2 \times 10^{-4}$ for both $G$ and $D$. The hyperparameter $\lambda$ of Consistency Regularization (CR) is set to $10$ as recommended~\cite{zhang2020CR}. All the models are run on 2 GPUs with a maximum of 250k training iterations on CIFAR-10 and 500k iterations on CIFAR-100.

For StyleGAN2, we use the official TensorFlow implementation\footnote{\url{https://github.com/NVlabs/stylegan2}} but include some changes to make it work better on CIFAR. The number of channels is $128$ at 32$\times$32 resolution and doubled at each coarser level with a maximum of $512$ channels. We set the half-life of the exponential moving average of the generator's weights to $1,000$k instead of $10$k images since it stabilizes the FID curve and leads to consistently better performance. We set $\gamma = 0.1$ instead of $10$ for the $R_1$ regularization, which significantly improves the baseline's performance under the 100\% data setting on CIFAR. The path length regularization and the lazy regularization are also disabled. The baseline model can already achieve the best FID and IS to our knowledge for unconditional generation on the CIFAR datasets. All StyleGAN2 models are trained on 4 GPUs with the default batch size ($32$) and a maximum training length of 25,000k images.

We apply \methodshort to BigGAN, CR-BigGAN, and StyleGAN2 without changes to the baseline settings. There are several things to note when applying \methodshort in conjunction with gradient penalties~\cite{gulrajani2017WGANGP} or CR~\cite{zhang2020CR}. The $R_1$ regularization penalizes the gradients of $D(\vv x)$ w.r.t.\@ the input $\vv x$. With \methodshort, the gradients of $D(T(\vv x))$ can be calculated w.r.t.\@ either $\vv x$ or $T(\vv x)$.
We choose to penalize the gradients of $D(T(\vv x))$ w.r.t.\@ $T(\vv x)$ for the CIFAR, FFHQ, and LSUN experiments since it slightly outperforms the other choice in practice; for the low-shot generation experiments, we penalize the gradients of $D(T(\vv x))$ w.r.t.\@ $\vv x$ instead from which we observe better diversity of the generated images.
As CR has already used image translation to calculate the consistency loss, we only apply \emph{Cutout} \methodshort on top of CR under the 100\% data setting. For the 10\% and 20\% data settings, we exploit stronger regularization by directly applying CR between $\vv x$ and $T(\vv x)$, i.e., before and after the \emph{Translation + Cutout} \methodshort.

We match the top performance for unconditional generation on CIFAR-100 as well as CIFAR-10 using only 20\% data (see \fig{fig:cifar100_results}). See \fig{fig:cifar10_1-analysis-IS} for the analysis of smaller models or stronger regularization in terms of IS. See \tab{tab:cifar10} and \tab{tab:cifar100} for quantitative results.

\subsection{Low-Shot Generation Experiments}
\lblsec{app:fewshot}
We compare our method to transfer learning algorithms using the FreezeD's codebase\footnote{\url{https://github.com/sangwoomo/FreezeD}} (for TransferGAN~\cite{wang2018TransferGAN}, Scale/shift~\cite{noguchi2019SB}, and FreezeD~\cite{mo2020FreezeD}) and the newly released MineGAN~\cite{wang2019MineGAN} code\footnote{\url{https://github.com/yaxingwang/MineGAN}}. All the models are transferred from a pre-trained StyleGAN model from the FFHQ dataset~\cite{karras2019StyleGAN} at 256$\times$256 resolution. %
FreezeD reports the best performance when freezing the first 4 layers of $D$~\cite{mo2020FreezeD}; when applying \methodshort to FreezeD, we only freeze the first 2 layers of $D$. All other hyperparameters are kept unchanged from the default settings. All the models are trained on 1 GPU with a maximum of 10k training iterations on our 100-shot datasets and 20k iterations on the AnimalFace~\cite{si2011AnimalFace} datasets.

When training the StyleGAN2 model from scratch, we use their default network configuration at 256$\times$256 resolution with an $R_1$ regularization $\gamma$ of $10$ but without the path length regularization and the lazy regularization. We use a smaller batch size of 16, which improves the performance of both the StyleGAN2 baseline and ours, compared to the default batch size of 32. All the models are trained on 4 GPUs with a maximum training length of 300k images on our 100-shot datasets and 500k images on the AnimalFace datasets.

\begin{figure}[t]
\begin{center}
\includegraphics[width=1.0\linewidth]{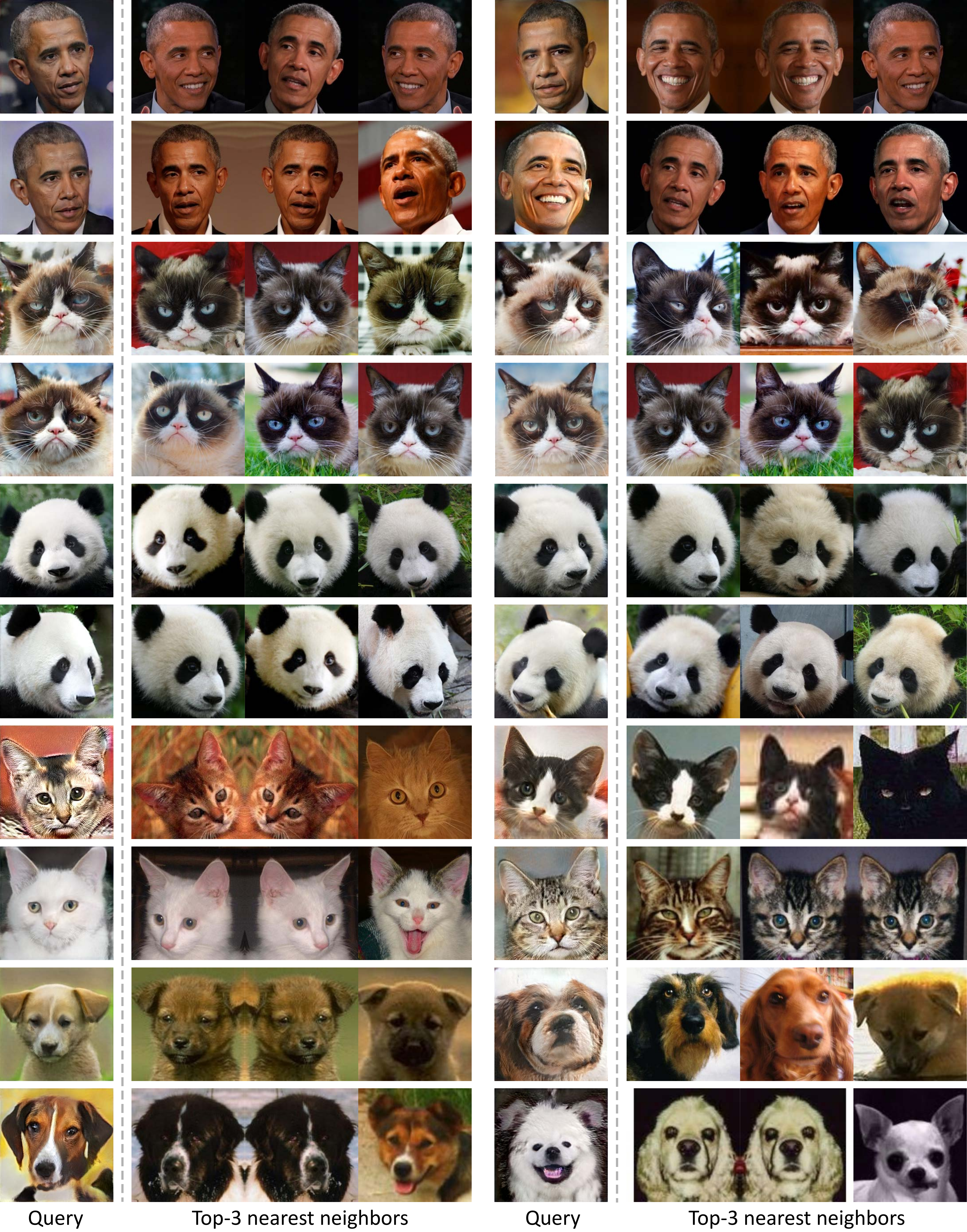}
\end{center}
\caption{\textbf{Nearest neighbors in pixel space} measured by the pixel-wise $L_1$ distance. Each query (on the left of the dashed lines) is a generated image of our method without pre-training (StyleGAN2 + \methodshort) on the 100-shot or AnimalFace generation datasets. Each nearest neighbor (on the right of the dashed lines) is an original image queried from the training set with horizontal flips. The generated images are different from the training set, indicating that our model does not simply memorize the training images or overfit even given small datasets.}
\label{fig:nearest_neighbors}
\end{figure}

\begin{figure}[t]
\begin{center}
\includegraphics[width=1.0\linewidth]{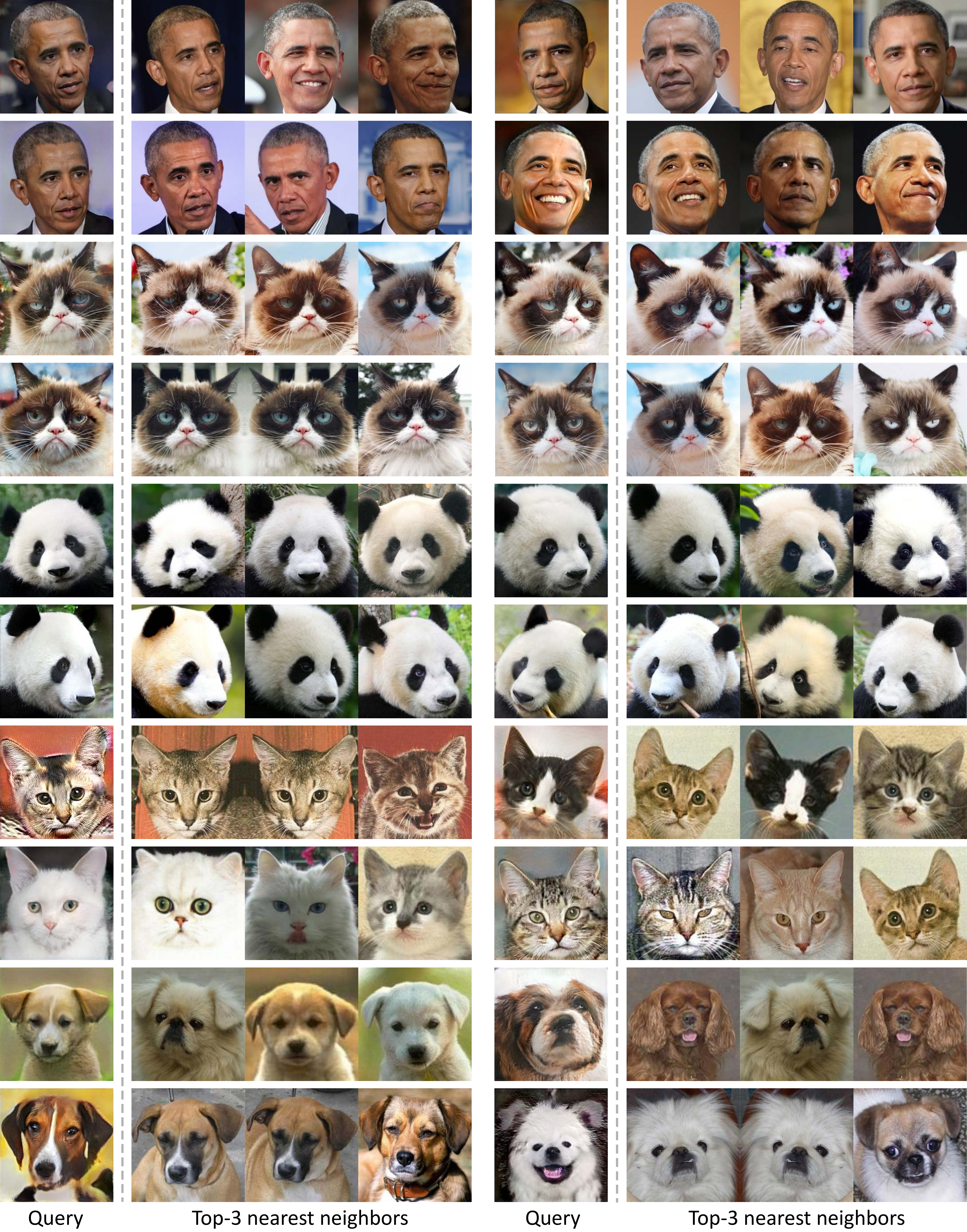}
\end{center}
\caption{\textbf{Nearest neighbors in feature space} measured by the Learned Perceptual Image Patch Similarity (LPIPS)~\cite{zhang2018LPIPS}. Each query (on the left of the dashed lines) is a generated image of our method without pre-training (StyleGAN2 + \methodshort) on the 100-shot or AnimalFace generation datasets. Each nearest neighbor (on the right of the dashed lines) is an original image queried from the training set with horizontal flips. The generated images are different from the training set, indicating that our model does not simply memorize the training images or overfit even given small datasets.}
\label{fig:nearest_neighbors_lpips}
\end{figure}

See \fig{fig:fewshot_interp_suppl} for the additional interpolation results, \fig{fig:nearest_neighbors} and \fig{fig:nearest_neighbors_lpips} for the nearest neighbor tests of our method without pre-training both in pixel space and in the LPIPIS feature space~\cite{zhang2018LPIPS}. See \figs{fig:few_shot-cat}-\ref{fig:few_shot-panda} for qualitative comparisons.

\section{Evaluation Metrics}

We measure FID and IS using the official Inception v3 model in TensorFlow for all the methods and datasets. Note that some papers using PyTorch implementations, including FreezeD~\cite{mo2020FreezeD}, report different numbers from the official TensorFlow implementation of FID and IS. On ImageNet, CIFAR-10, and CIFAR-100, we inherit the setting from the Compare GAN codebase that the number of samples of generated images equals the number of real images in the validation set, and the validation set is used as the reference distribution for FID calculation. For the low-shot generation experiments, we sample 5k generated images and we use the training set as the reference distribution. For the FFHQ and LSUN experiments, we use the same evaluation setting as ADA~\cite{karras2020ADA}.

\section{100-Shot Generation Benchmark}

We collect the 100-shot datasets from the Internet. We then manually filter and crop each image as a pre-processing step. The full datasets are available \href{https://hanlab.mit.edu/projects/data-efficient-gans/datasets/}{here}.

\section{Changelog}
\begin{figure}[t]
\begin{center}
\includegraphics[width=1.0\linewidth]{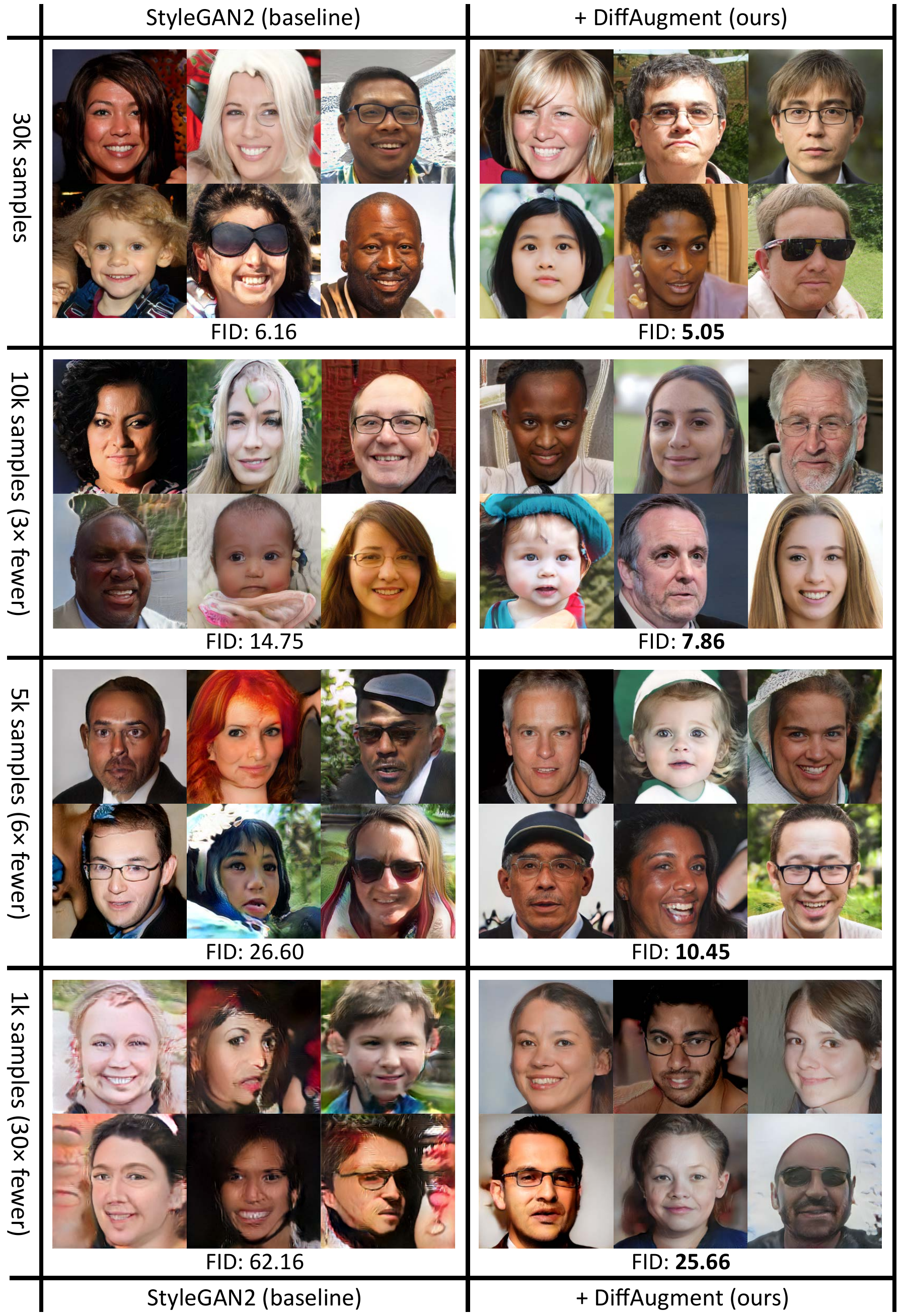}
\end{center}
\caption{Qualitative comparison on \textbf{FFHQ} at 256$\times$256 resolution with 1k, 5k, 10k, and 30k training images. Our method consistently outperforms the StyleGAN2 baselines~\cite{karras2019StyleGAN2} under different data percentage settings. }
\label{fig:ffhq_comparison}
\end{figure}

\begin{figure}[t]
\begin{center}
\includegraphics[width=1.0\linewidth]{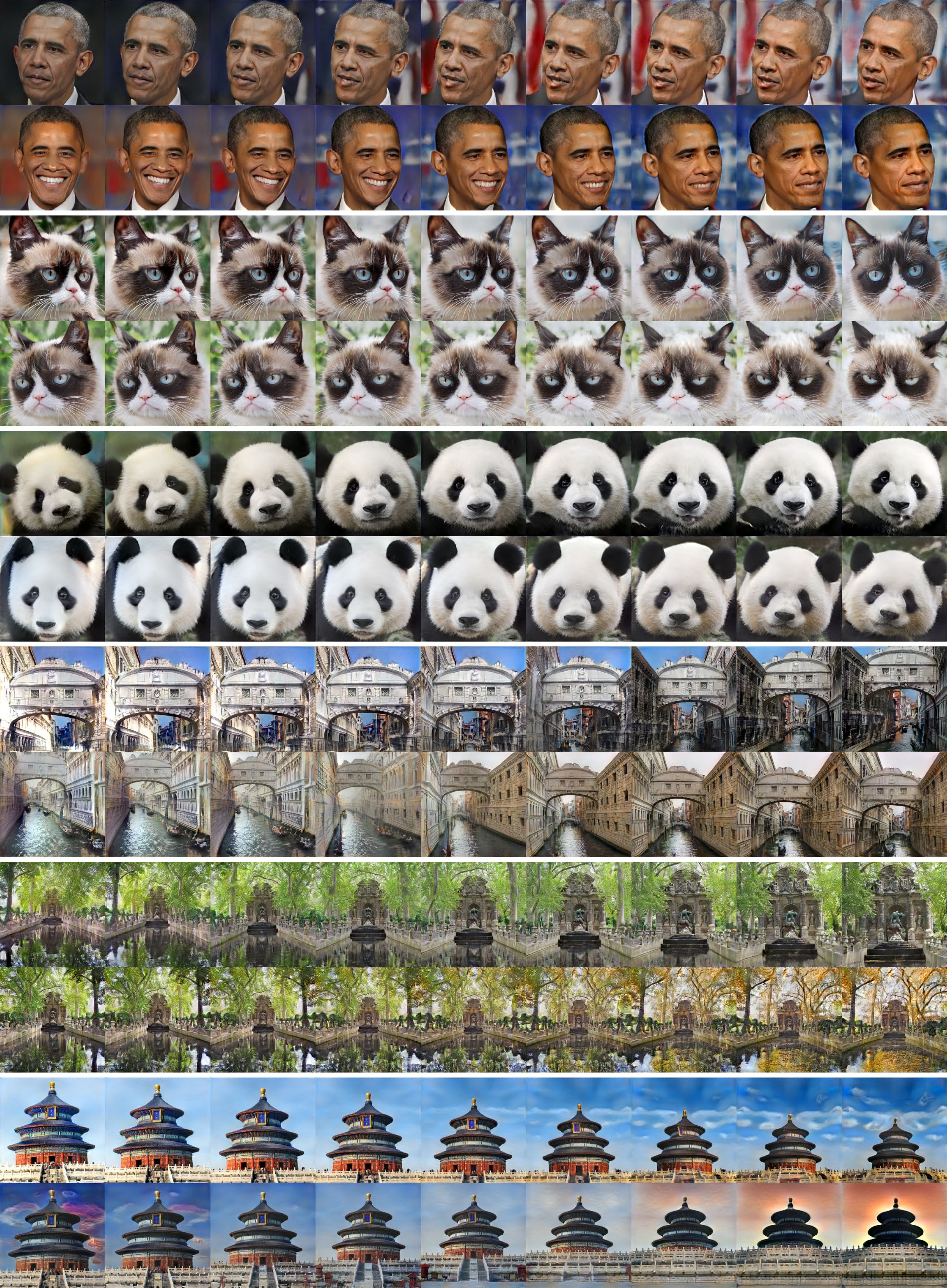}
\end{center}
\caption{\textbf{Style space interpolation} of our method on the 100-shot Obama, grumpy cat, panda, the Bridge of Sighs, the Medici Fountain, and the Temple of Heaven datasets without pre-training. The smooth interpolation results suggest little overfitting of our method even given small datasets.}
\label{fig:fewshot_interp_suppl}
\end{figure}

\myparagraph{v1} Initial preprint release.
\myparagraph{v2} (a) Add StyleGAN2 results on FFHQ (\reftbl{ffhq} and \reffig{ffhq_comparison}). (b) For low-shot generation, we rerun the StyleGAN2 and StyleGAN + \methodshort models with a batch size of 16 (\reftbl{few_shot}). Both our method and the baseline are improved with a batch size of 16 over 32 (used in v1). (c) Add interpolation results on the additional 100-shot landmark datasets (\reffig{fewshot_interp_suppl}). (d) Update the CR-BigGAN notes in \refapp{app:imagenet}.

\begin{figure}[t]
\begin{center}
\includegraphics[width=1.0\linewidth]{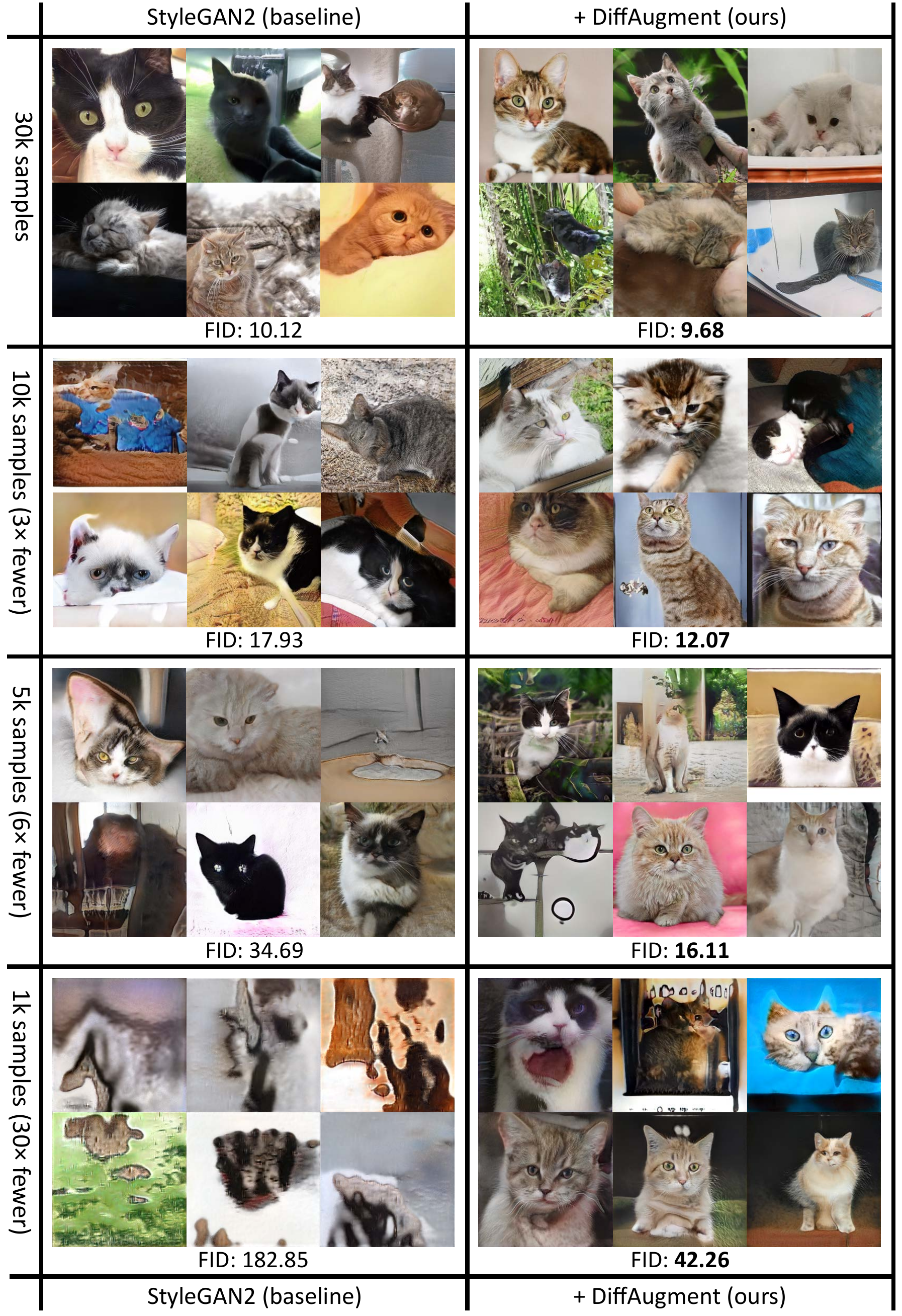}
\end{center}
\caption{Qualitative comparison on \textbf{LSUN-cat} at 256$\times$256 resolution with 1k, 5k, 10k, and 30k training images. Our method consistently outperforms the StyleGAN2 baselines~\cite{karras2019StyleGAN2} under different data percentage settings. }
\label{fig:lsun_cat_comparison}
\end{figure}

\begin{figure}[t]
\begin{center}
\includegraphics[width=0.95\linewidth]{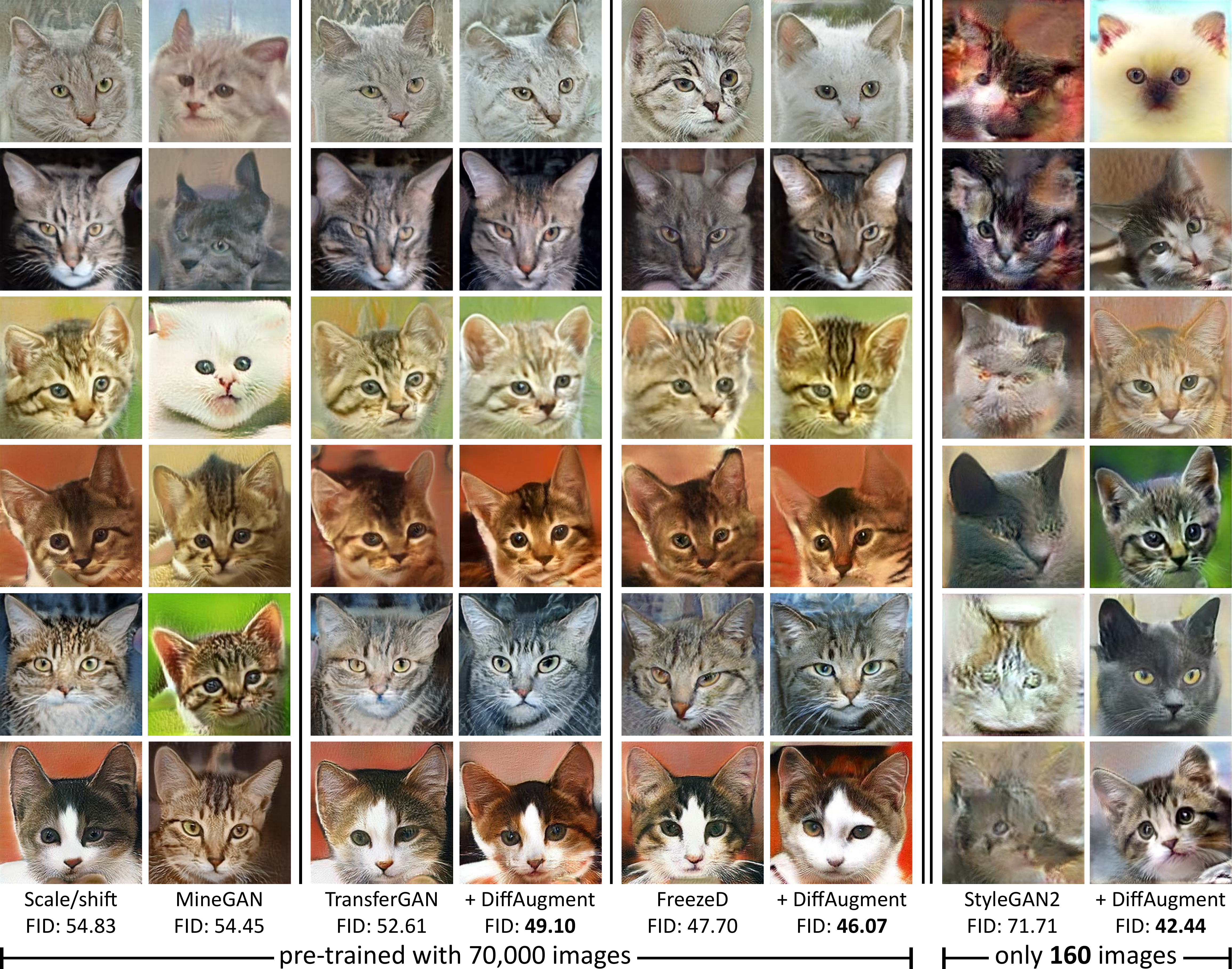}
\end{center}
\vspace{-8pt}
\caption{Qualitative comparison on the \textbf{AnimalFace-cat}~\cite{si2011AnimalFace} dataset.}
\label{fig:few_shot-cat}
\vspace{-16pt}
\end{figure}

\begin{figure}[t]
\begin{center}
\includegraphics[width=0.95\linewidth]{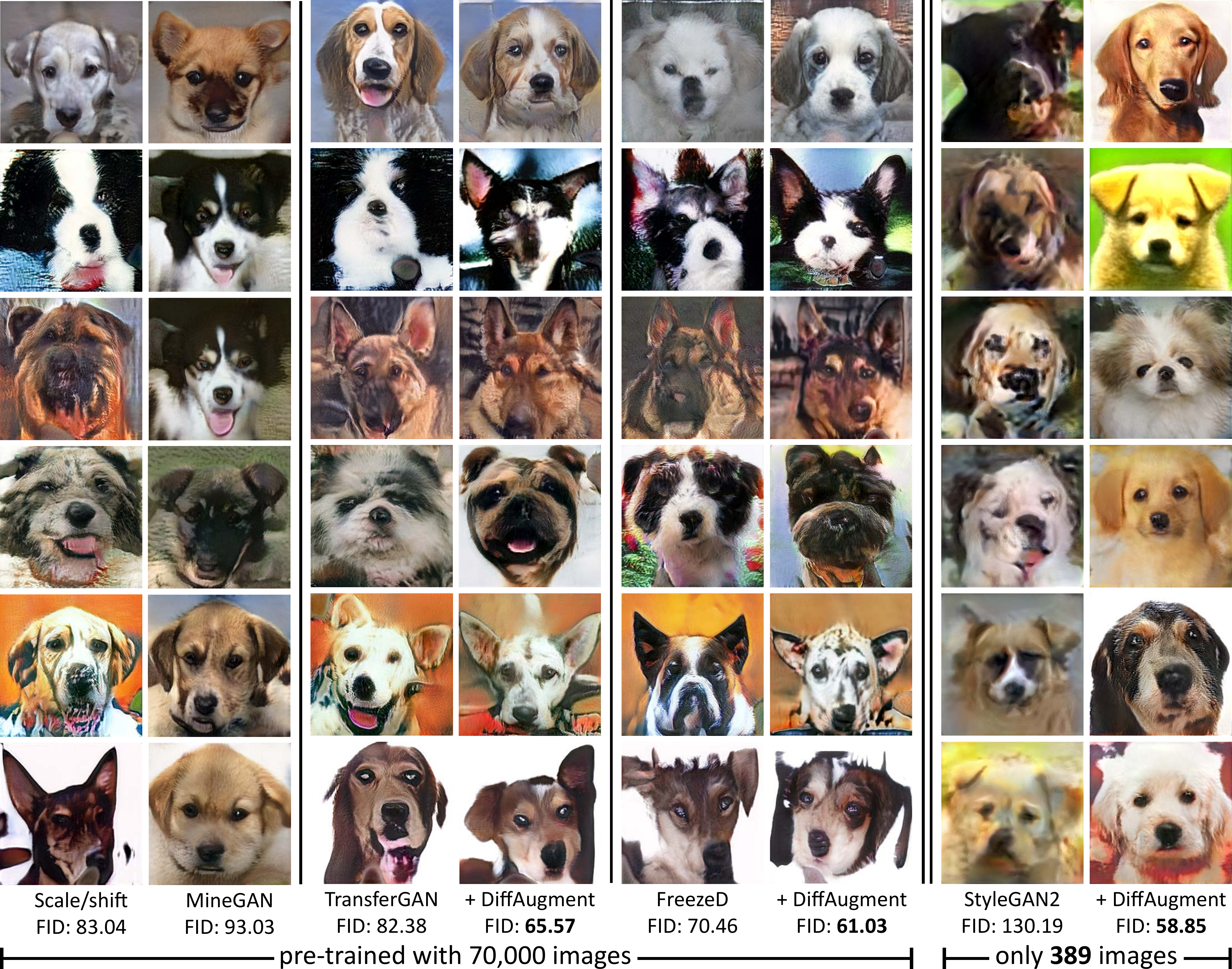}
\end{center}
\vspace{-8pt}
\caption{Qualitative comparison on the \textbf{AnimalFace-dog}~\cite{si2011AnimalFace} dataset.}
\label{fig:few_shot-dog}
\vspace{-16pt}
\end{figure}

\begin{figure}[t]
\begin{center}
\includegraphics[width=0.9\linewidth]{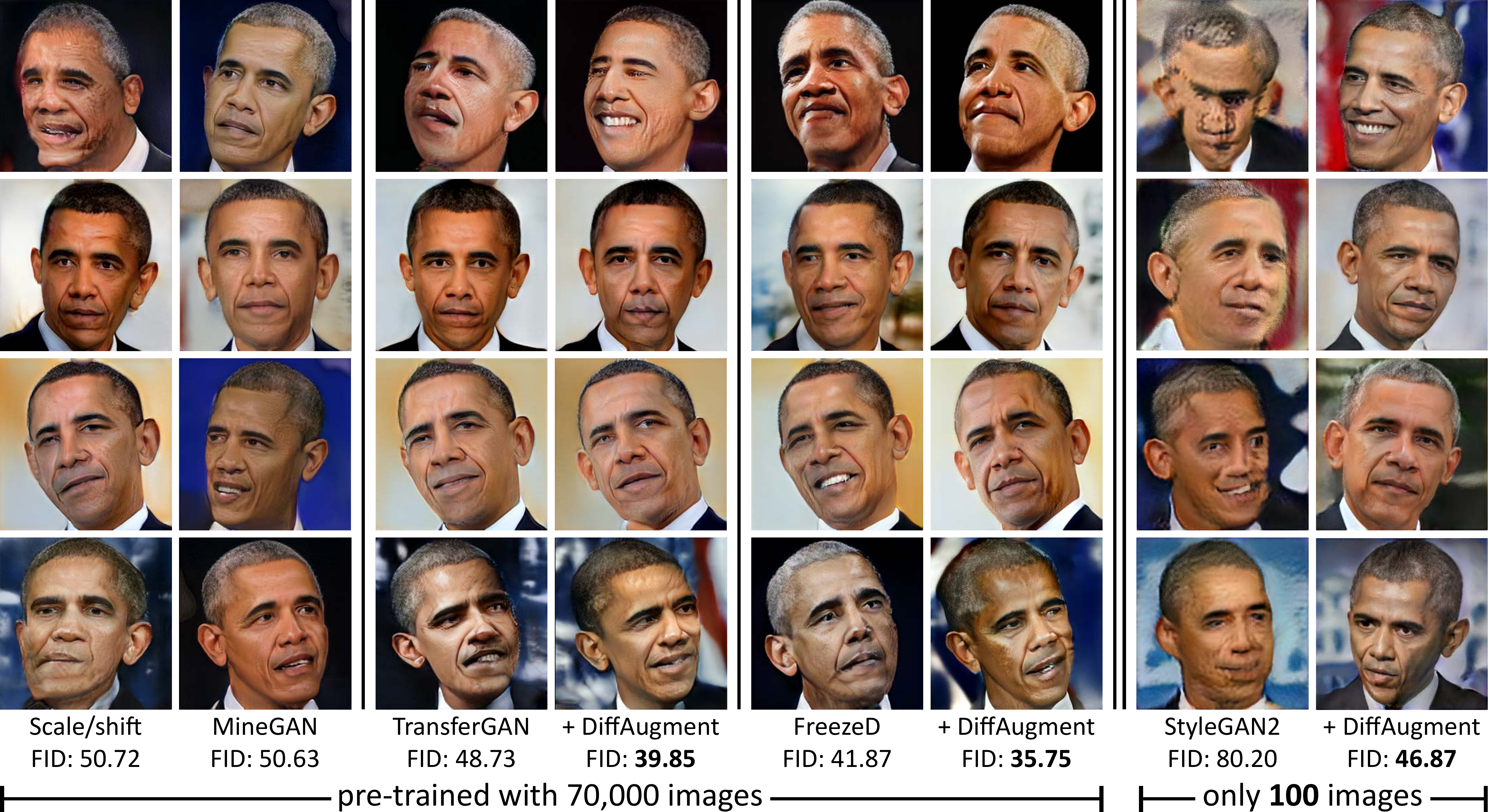}
\end{center}
\vspace{-8pt}
\caption{Qualitative comparison on the \textbf{100-shot Obama} dataset.}
\label{fig:few_shot-obama}
\vspace{-16pt}
\end{figure}

\begin{figure}[t]
\begin{center}
\includegraphics[width=0.9\linewidth]{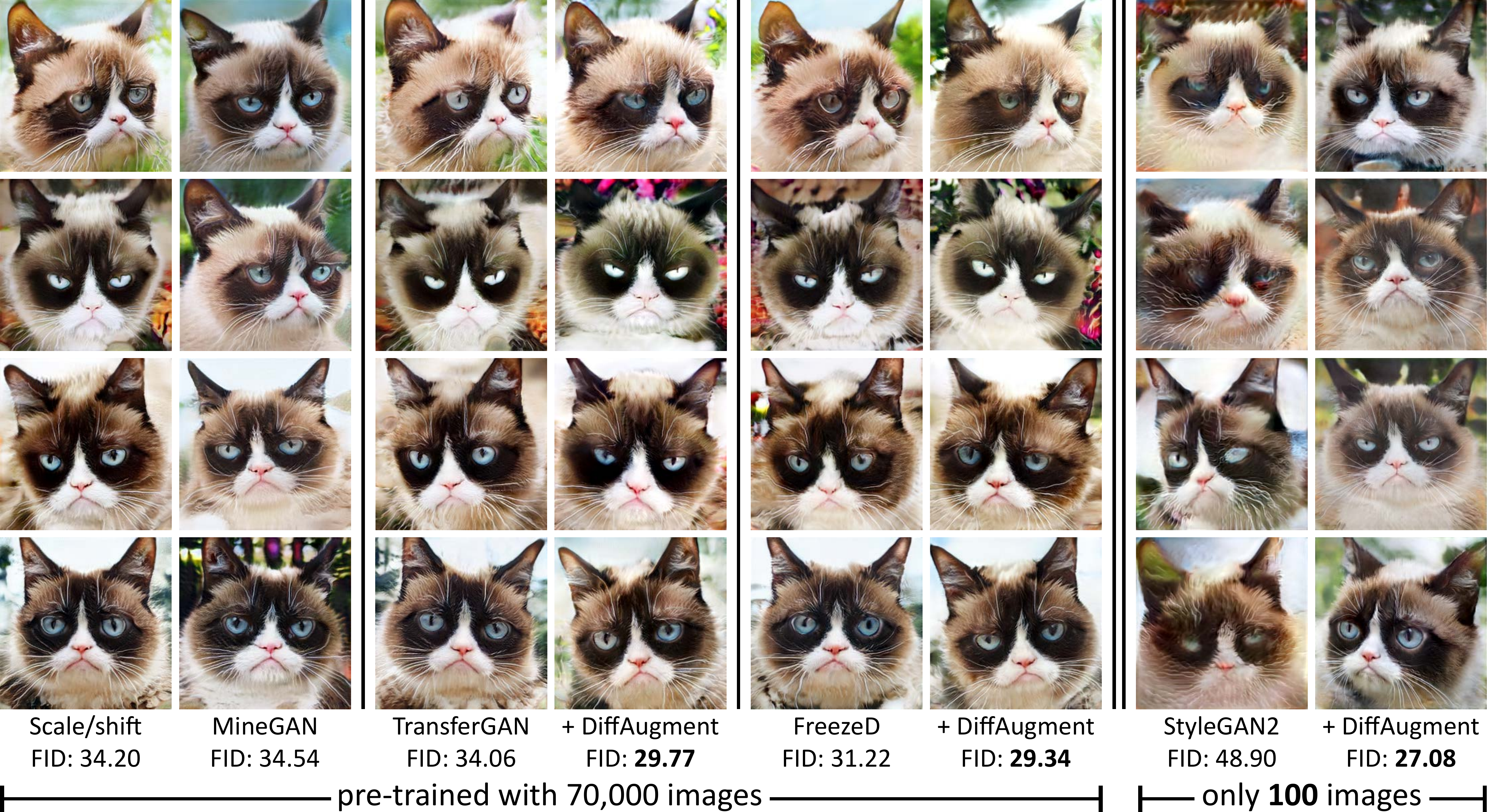}
\end{center}
\vspace{-8pt}
\caption{Qualitative comparison on the \textbf{100-shot grumpy cat} dataset.}
\label{fig:few_shot-grumpy_cat}
\vspace{-16pt}
\end{figure}

\begin{figure}[t]
\begin{center}
\includegraphics[width=0.9\linewidth]{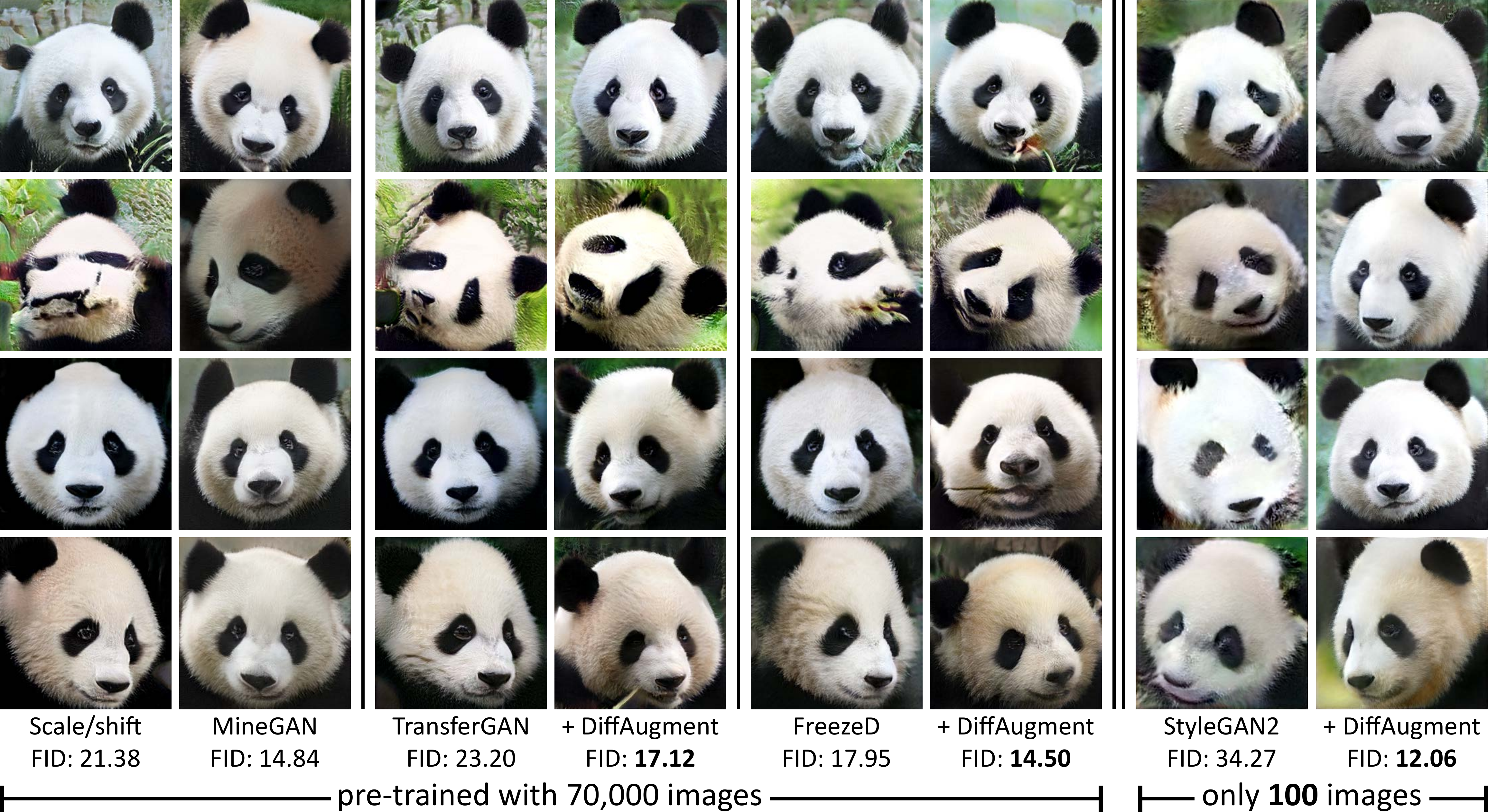}
\end{center}
\vspace{-8pt}
\caption{Qualitative comparison on the \textbf{100-shot panda} dataset.}
\label{fig:few_shot-panda}
\vspace{-16pt}
\end{figure}

\myparagraph{v3} \added{(a) Add StyleGAN2 results on the LSUN-Cat dataset (\reftbl{ffhq} and \reffig{lsun_cat_comparison}). (b) Rerun the MineGAN models using their newly released code (\reftbl{few_shot}), and add qualitative samples (\figs{fig:few_shot-cat}-\ref{fig:few_shot-panda}). (c) Analyze additional choices of \methodshort (\reffig{aug-analysis})}.

\myparagraph{v4} \added{Add notes in \reftbl{cifar} regarding a different evaluation protocal for FID calculation as used in concurrent works~\cite{karras2020ADA,ho2020denoising}.}

\clearpage

\end{appendices}

\end{document}